\documentclass{article}

\PassOptionsToPackage{square,numbers}{natbib}
\usepackage{ml_arxiv}

\usepackage{graphicx}
\usepackage{hyperref}
\usepackage{booktabs}
\usepackage{xcolor}
\usepackage{subcaption}
\usepackage{amsmath}
\usepackage{amssymb}
\usepackage{bm}

\title{AB-UPT for Automotive and Aerospace Applications}

\author{
Benedikt Alkin$^{1}$ \quad
Richard Kurle$^{1}$ \quad
Louis Serrano$^{1}$ \quad \\
\textbf{Dennis Just}$^{1}$    \quad
\textbf{Johannes Brandstetter}$^{1,2}$\\  
$~^{1}$EMMI AI, Linz, Austria\\
$~^{2}$ELLIS Unit Linz, Institute for Machine Learning, JKU Linz, Austria\\
}

\date{September 2025}

\begin{document}

\maketitle

\begin{abstract}
    The recently proposed Anchored-Branched Universal Physics Transformers (AB-UPT) shows strong capabilities to replicate automotive computational fluid dynamics simulations requiring orders of magnitudes less compute than traditional numerical solvers. In this technical report, we add two new datasets to the body of empirically evaluated use-cases of AB-UPT, combining high-quality data generation with state-of-the-art neural surrogates. Both datasets were generated with the Luminary Cloud platform containing automotives (SHIFT-SUV) and aircrafts (SHIFT-Wing).
    We start by detailing the data generation. Next, we show favorable performances of AB-UPT against previous state-of-the-art transformer-based baselines on both datasets, followed by extensive qualitative and quantitative evaluations of our best AB-UPT model. AB-UPT shows strong performances across the board. Notably, it obtains near perfect prediction of integrated aerodynamic forces within seconds from a simple isotopically tesselate geometry representation and is trainable within a day on a single GPU, paving the way for industry-scale applications.
\end{abstract}

\section{Introduction}

Accurate, fast neural surrogates for computational fluid dynamics (CFD) have gathered significant interest from industry and engineers due to the computational costs of classical CFD simulations, where their neural surrogate is typically orders of magnitudes faster. Recent advances in architecture and training methodology has enabled training accurate Transformer architectures~\cite{vaswani2017transformer,dosovitsky2021vit,alkin2024upt,alkin2024neuraldem} on CFD meshes containing hundreds of millions of mesh cells by only using a fraction of the CFD mesh cell centers for training while allowing efficient inference~\cite{alkin2025abupt}. This architecture, called Anchored-Branched Universal Physics Transformers (AB-UPT), showed accurate modeling of surface and volume fields as well as high correlation with integrated quantities such as drag coefficient on high-fidelity automotive CFD simulations. 

In this technical report we train and evaluate AB-UPT models on two recently introduced datasets from the Luminary Cloud SHIFT program. Namely, SHIFT-SUV, a dataset of automotive CFD simulations where car-design varies between samples, and SHIFT-Wing, a dataset of airplane CFD simulations with different plane-designs, angle-of-attack regimes, and mach numbers. We evaluate our AB-UPT surrogate model for practical engineering use-cases by showing quantitative per-field metrics, predictive accuracy of integrated drag and lift coefficients, training and inference costs, as well as a best and worst case analysis. Our results show good agreement on simulations from the hold-out testset enabling order of magnitude faster CFD simulations with high accuracy such that the surrogate models are relevant to practitioners for, e.g., design optimization use-cases.

\section{Preliminaries}

This work leverages the AB-UPT architecture~\cite{alkin2025abupt} where we do not introduce significant changes to architecture or training procedure. Therefore, we refer to~\cite{alkin2025abupt} for more detailed motivation, background, architecture description and related work. Here we describe only the most important architectural and training details as well as dataset specific background.

\subsection{AI surrogate modeling for external aerodynamics}

In general, a surrogate~\citep{forrester2008engineering}, or surrogate model, is a simpler model used to approximate a more complex, computationally demanding system like a simulation or physical experiment.

There has been a recent surge in the use of transformers to create these surrogate models, building on their success in scientific fields like protein folding~\citep{jumper_highly_2021,abramson_accurate_2024} and weather modeling~\citep{bi_accurate_2023,bodnar_aurora_2024}. These transformer-based surrogates are designed to effectively integrate information across different spatial locations and scales by leveraging attention mechanisms~\citep{vaswani2017transformer}. Transformer-based surrogates extend the neural operator paradigm~\citep{Lu:19, Lu:21, Li:20graph, Li:20,Kovachki:21} by incorporating self-attention, cross-attention, or perceiver blocks~\citep{jaegle2021perceiver,jaegle2021perceiverIO}. Specific examples of these models include OFormer~\citep{li2022transformer}, Transolver~\citep{wu2024transolver}, and (AB-)UPT~\citep{alkin2024upt,alkin2025abupt}.

\subsection{Anchored-Branched Universal Physics Transformer (AB-UPT)} AB-UPT~\cite{alkin2025abupt} is a transformer-based model to tackle large-scale physical simulations. It treats the input mesh as a pointcloud and maps it to corresponding surface and volume fields where surface and volume are logically separated into different branches with regular interactions throughout the forward pass. AB-UPT leverages the expressive vanilla attention mechanism~\cite{vaswani2017transformer} where its quadratic cost is kept in check via training on a drastically reduced set of input points, so-called \textit{anchor points}, together with a mechanism to efficiently infer the full set of points during inference via cross-attending only to anchor points.

The original work showed strong performances coupled with high efficiency on CFD simulations with up to 140 million volumetric cells that were simulated with a hybrid RANS-LES (HRLES)~\citep{spalart2006new,chaouat2017state,heinz2020review,ashton2022hlpw} approach, which is a routine approach in the automotive industry~\cite{hupertz2022towards,ashton2024drivaerml}, outperforming other models like Transolver~\cite{wu2024transolver}. In this report, we show that AB-UPT outperforms similar models on the SHIFT-SUV (containing car simulations) and SHIFT-Wing (containing airplane simulations) datasets. Additionally, we investigate the performance of AB-UPT w.r.t.\ the practical usability.

We only make one small adjustment to the AB-UPT model, namely that we do not share parameters between branches as we found this to perform slightly better without increasing runtime.

\subsection{Aerodynamic forces}
\label{sec:aero_forces}

Aerodynamic forces quantify interesting properties of the car/airplane design such as drag and lift forces which are the result of an integration of the surface pressure $p_s$ and shear stress $\bm{\tau}_w$, i.e., force per unit area exerted by the fluid on the surface, acting tangential (parallel) to the surface.
Aerodynamic forces are obtained from the force acting on an object in an airflow, which is given by
\begin{align}
    \bm{F} = \oint_S \bigg( -(p_s-p_\infty) \bm{n} + \bm{\tau}_w \bigg) dS \ ,
\end{align}
where $p_s$ is the surface pressure, $p_\infty$ the free stream pressure, $\bm{n}$ the surface normal vector, and $\bm{\tau}_w$ the shear stress contribution.
For comparability between designs, dimensionless numbers as drag and lift coefficients
\begin{equation}
    C_\text{d} =   \frac{2\, \bm{F} \cdot \bm{e}_{\text{drag}}}{\rho\, v^2 A_\text{ref}}, \; C_\text{l} =   \frac{2\, \bm{F} \cdot \bm{e}_\text{lift}}{\rho\, v^2 A_\text{ref}}
    \label{eq:drag_and_lift_coefficient}
\end{equation}
are often used in engineering~\citep{ashton2024drivaerml}, where $\bm{e}_{\text{drag}}$ is a unit vector into the free stream direction, $\bm{e}_{\text{lift}}$ a unit vector into the lift direction perpendicular to the free stream direction, $\rho$ the density, $v$ the magnitude of the free stream velocity, and $A_{\text{ref}}$ a characteristic reference area.
Further, the acting force $\bm{F}$ can be converted into dimensionless numbers as drag and lift forces

\begin{equation}
    \bm{F}_\text{drag} =   \bm{F} \cdot \bm{e}_{\text{drag}}, \; \bm{F}_\text{lift} =   \bm{F} \cdot \bm{e}_\text{lift} \ .
    \label{eq:drag_and_lift_force}
\end{equation}



When simulating an airplane in flight, the angle of attack $\alpha$ is a common boundary condition which changes free stream $\bm{e}_\text{drag}$ and lift direction $\bm{e}_\text{lift}$ as given by the unit velocity vector $\bm{u}_\infty$ as
\begin{equation}
    \bm{u}_\infty = \begin{bmatrix} \cos(\alpha) \\ 0 \\ \sin(\alpha) \end{bmatrix}, \;
    \bm{e}_\text{drag} = \frac{\bm{u}_\infty}{\|\bm{u}_\infty\|_2}  , 
    \; \bm{e}_\text{lift} = \bm{e}_\text{drag} \times \begin{bmatrix} 0 \\ 1 \\ 0 \end{bmatrix} \ , 
    \label{eq:drag_and_lift_direction}
\end{equation}

where $\|.\|_2$ is the Euclidian norm. For $\alpha=0$ these reduce to $\bm{e}_\text{drag} = (1, 0, 0)$ and $\bm{e}_\text{lift} = (0, 0, 1)$.

We choose to compare aerodynamic forces of Equation~\eqref{eq:drag_and_lift_force} instead of normalizing them to represent a coefficient in order to make comparison with a relative error (as outlined in the next section) easier. Drag and lift coefficients can be close to zero which leads to unintuitive relative error values. For the datasets considered in this work, comparing aerodynamic forces does not run into this issue as they have magnitudes larger than 1.

\section{Experiments}

\subsection{Model configuration and baseline models}


We evaluate the performance of AB-UPT on the SHIFT-SUV and SHIFT-Wing datasets by benchmarking it against several baselines. 
These include other strong transformer-based surrogates (Transolver, standard Transformer) and DoMINO \cite{ranade2025domino}; the latter is a prominent point-cloud-based neural operator for 3D CFD modeling whose architecture deviates from the state-of-the-art attention mechanism of transformers.
We provide various practical evaluation metrics and visualizations to showcase the usefulness of these models. 
For the transformer models, we use the same hyperparameter configurations as in the AB-UPT paper \cite{alkin2025abupt}:
each model consists of 12 transformer blocks with a hidden dimension of 192, 3-headed attention, and an MLP that expands to 4 times the hidden dimension using a GELU activation function. 
Transolver and Transformer models use full self-attention between surface and volume tokens, resulting in approximately 5.5M parameters. 
In contrast, AB-UPT interleaves self- and cross-attention within and between the respective domains, where the weights in these two branches can be (partially) shared or separated. 
In this work, we use separate weights, which roughly doubles the number of parameters but keeps the same computational cost.
For the DoMINO model, we use the official implementation from the physics-nemo repository with the hyperparameter configuration from \citet{ranade2025domino}, resulting in a model with 19.7M parameters. Beyond raw geometric features, DoMINO incorporates preprocessed surface normal vectors, surface areas, and signed distance functions as auxiliary inputs, which enhance the model's ability to capture the geometric characteristics of the aerodynamic domain. The training loss combines volume, surface, and area-weighted components; specifically, the surface loss is scaled 5× relative to the volume loss to emphasize surface prediction accuracy. 
On the SHIFT-SUV dataset, we also evaluated AB-UPT with a hidden dimension of 384 using 6 attention heads. 
For the SHIFT-Wing dataset, where the angle of attack (AoA) is varied, we use the method proposed in \citet{peebles22dit} to condition each transformer block on the AoA. 
An overview of the dataset properties is outlined in Table~\ref{table:datasets}.

\begin{table}[h!]
\centering
\begin{tabular}{lcccccc}
& \multicolumn{2}{c}{{\#Points}} & \multicolumn{2}{c}{{Properties}} \\
\cmidrule(rl){2-3} \cmidrule(rl){4-5} 
Dataset & Surface & Volume & \#Simulations & Variability \\
\midrule
SUV & 3M & 45M & 3992 & Geometry~\cite{heft2012drivaer,zhang2019aerosuv}, physical model scale $\in \{1, \frac{1}{4} \}$ \\
WING & 3M & 6M & 1698 & Geometry~\cite{nasa_crm}, AoA $\in [0^\circ, 4^\circ] $, Mach $\in \{0.5, 0.85\} $ \\
\end{tabular}
\caption{Both datasets contain pressure $p_s$ and shear stress $\bm{\tau}$ as surface variables as well as pressure $p_v$ and velocity $\bm{u}$ as volume variables. AoA denotes the angle-of-attack. Geometries are varied via parametric models. As AB-UPT is a point-based model, we use the centers of the CFD mesh cells as inputs.  }
\label{table:datasets}
\end{table}


\subsection{Luminary SHIFT-SUV}

\subsubsection{Dataset description}
\label{sec:dataset_suv}

SHIFT-SUV is an open-source aerodynamics database of high-fidelity external aerodynamics simulations developed by Luminary Cloud. The dataset (which continues to increase in size) currently includes over four thousand transient delayed detached eddy simulations (DDES) of geometric variants of the AeroSUV vehicle platform \citep{zhang2019aerosuv}. The AeroSUV platform, developed by FKFS (Forschungsinstitut für Kraftfahrwesen und Fahrzeugmotoren Stuttgart), shares inspiration and geometry characteristics from the well known DrivAer platform popular in sedan automotive aerodynamics~\cite{heft2012drivaer}. SHIFT-SUV is provided with the CC-BY-NC license and can be downloaded at no cost through HuggingFace~\cite{shiftsuv2025}.

\begin{figure}[h!]
\centering
\begin{subfigure}{0.5\textwidth}
    \centering
    \includegraphics[width=0.99\linewidth]{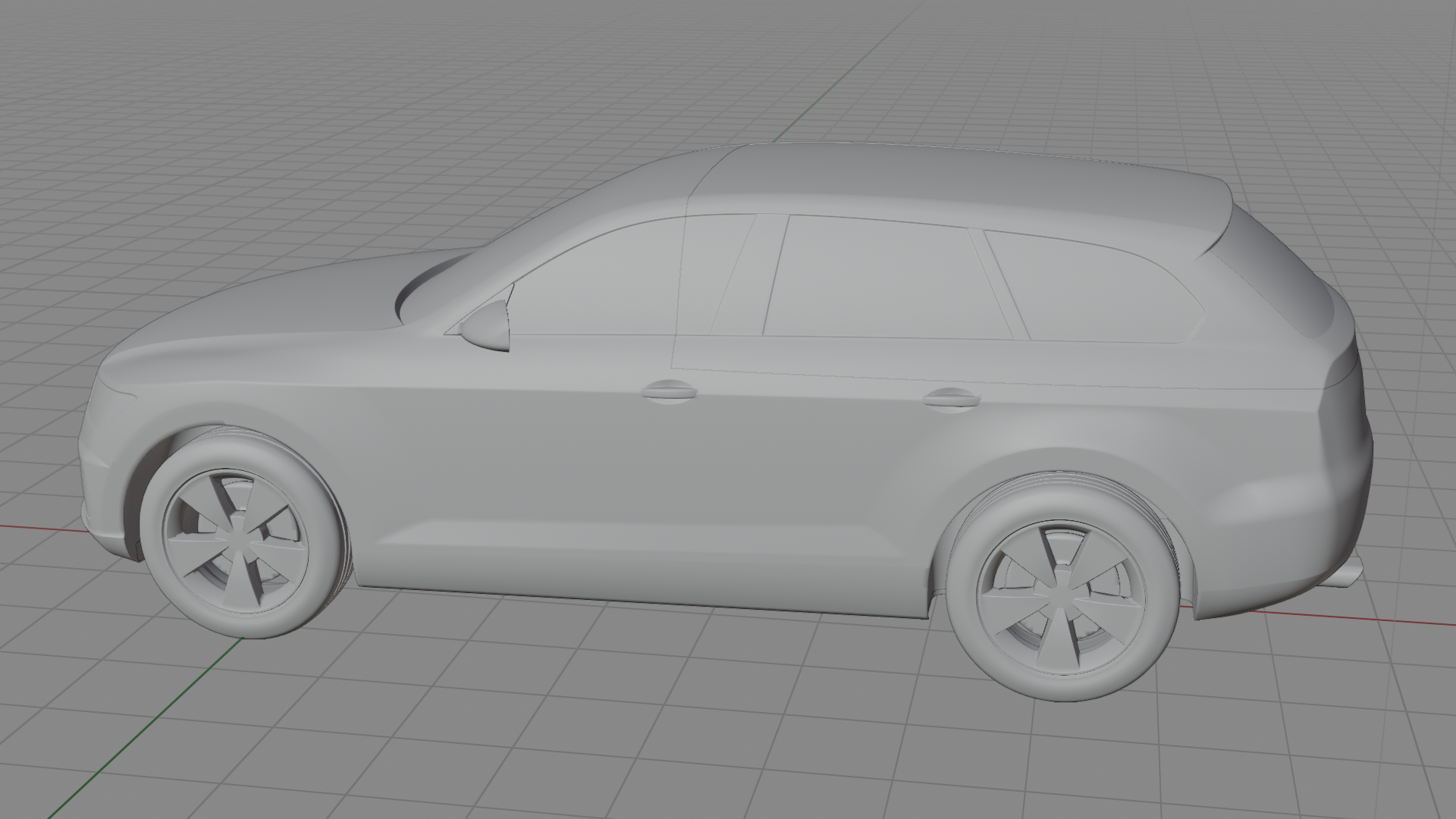}
\end{subfigure}%
\hfill
\begin{subfigure}{0.5\textwidth}
    \centering
    \includegraphics[width=0.99\linewidth]{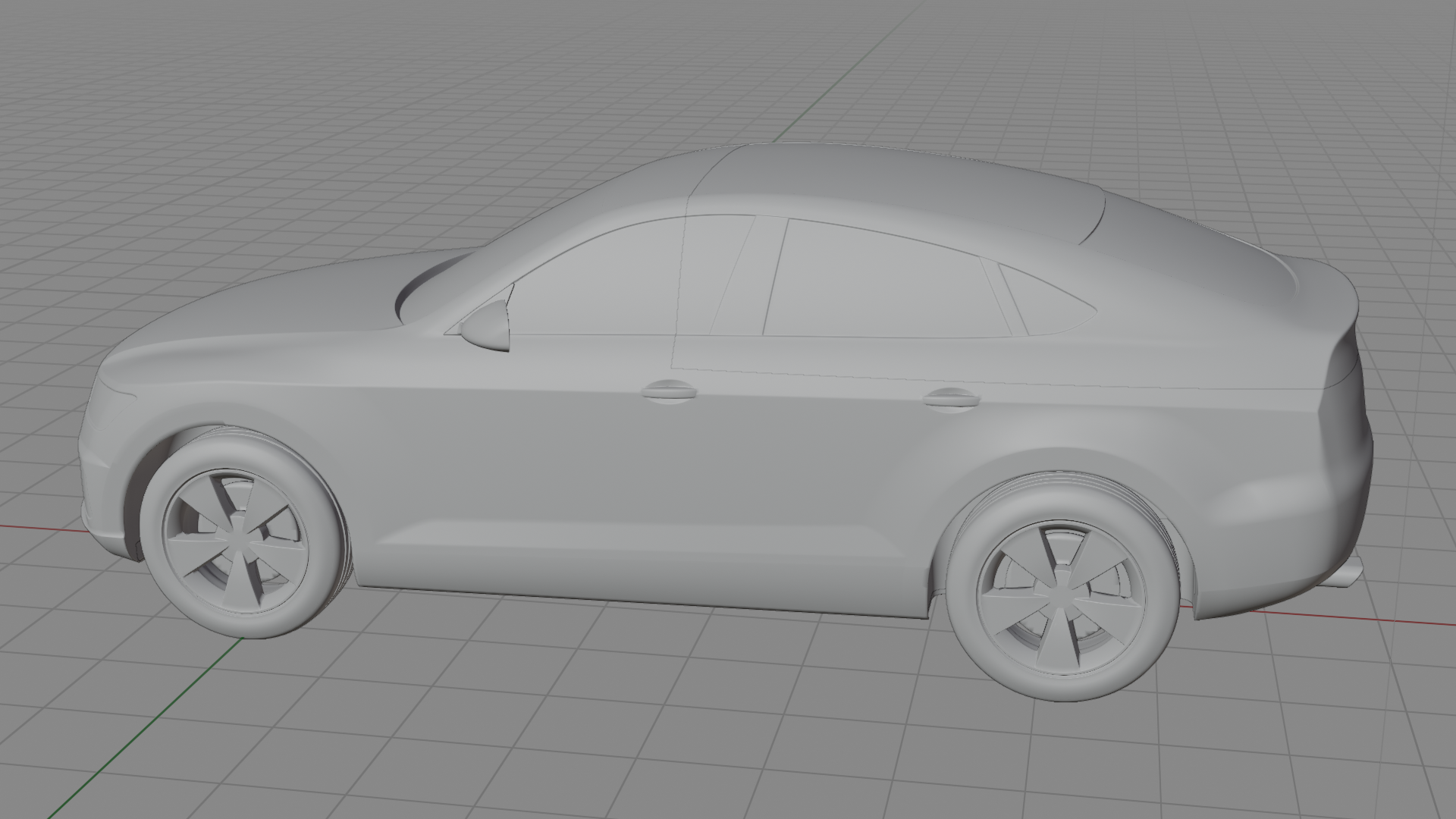}
\end{subfigure}
\caption{ Estate (left) and Fastback (right) configurations of the reference AeroSUV platform.
}
\label{fig:aerosuv_reference}
\end{figure}

\begin{figure}[h!]
\centering
\begin{subfigure}{0.8\textwidth}
    \centering
    \includegraphics[width=0.99\linewidth]{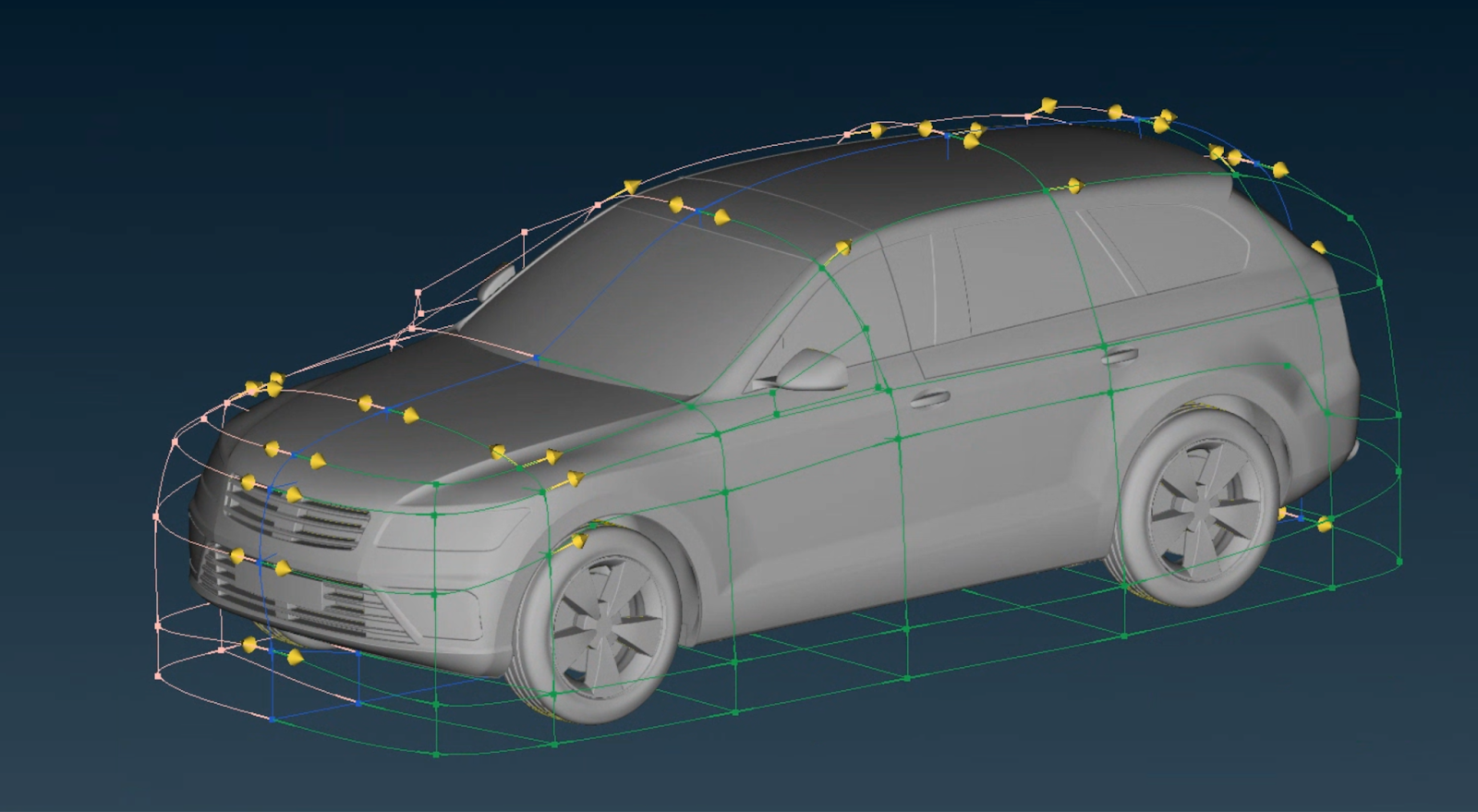}
\end{subfigure}
\caption{ The deformation cage used to create the geometry variants constructed around the Estate configuration.
}
\label{fig:aerosuv_cage}
\end{figure}

Geometry variants are produced using morphing boxes defined about the reference configurations using the ANSA software from BETA-CAE Systems, shown in Figure~\ref{fig:aerosuv_cage}. The specific parameterization and their extents were informed through partnership with Honda Motor Company, and are summarized in Table~\ref{table:suv_morph_params}. The deformations are described in design parameters familiar to vehicle stylists, and notably are not mutually orthogonal, e.g., multiple parameters may modify the same cage control points. To ensure consistency of geometry that should remain fixed across all variants, e.g., wheels, tires, and suspension components, these geometries regions are not influenced by the morphing approach. Based on input from Honda, the initial dataset explores only variations of these geometry parameters rather than also changing the boundary conditions.

\begin{figure}[h!]
\centering
\begin{subfigure}{0.5\textwidth}
    \centering
    \includegraphics[width=0.99\linewidth]{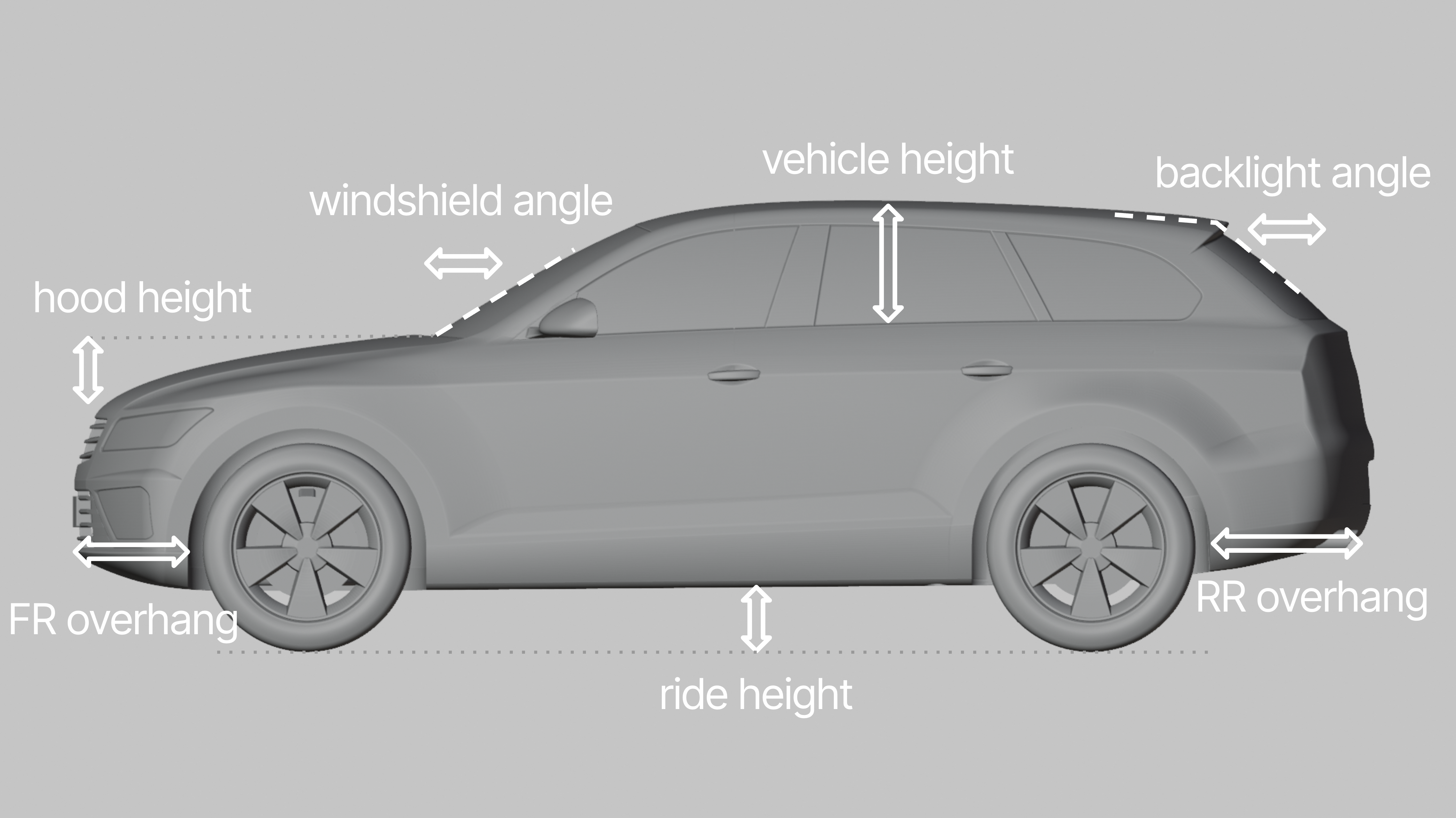}
\end{subfigure}%
\hfill
\begin{subfigure}{0.5\textwidth}
    \centering
    \includegraphics[width=0.99\linewidth]{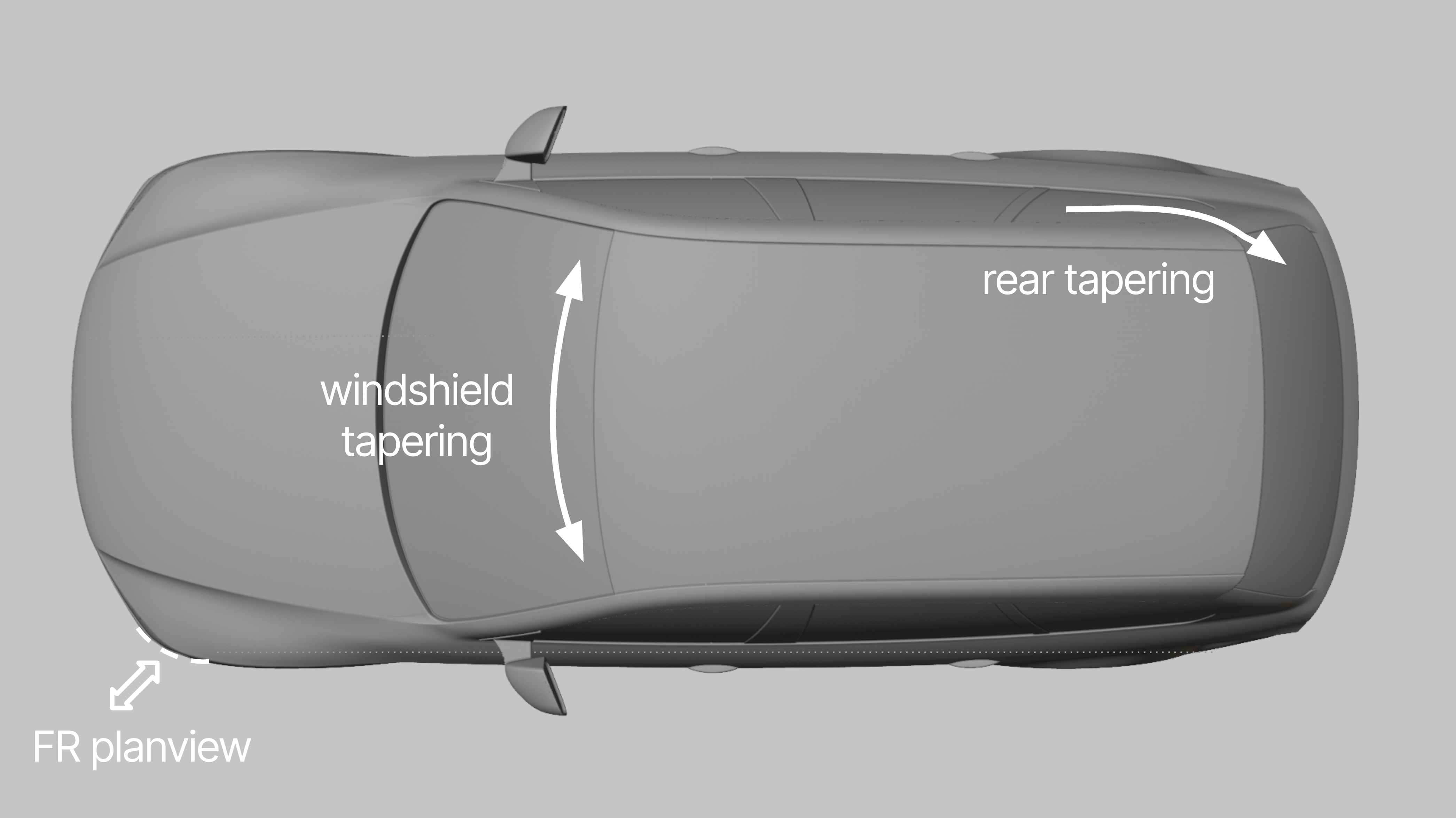}
\end{subfigure}
\caption{ Diagrams visualizing the deformation parameters described in Table~\ref{table:suv_morph_params}. The parameters in the right image modify the vehicle symmetrically about the centerline (left and right sides equally).
}
\label{fig:aerosuv_params_diagrams}
\end{figure}

\begin{table}[h!]
\centering
\begin{tabular}{lccc}
Deformation parameter & Min [mm] & Max [mm] \\
\midrule
Hood Height & -50 & +50 \\
FR Overhang & -150 & +150 \\
Windshield Angle & -150 & +100 \\
Vehicle Height & -150 & +150 \\
Ride Height & -30 & +30 \\
Backlight Angle & -100 & +200 \\
RR Overhang & -150 & +100 \\
FR Planview & -75 & +75 \\
Windshield Tapering & -100 & +100 \\
Rear Tapering & -90 & +70 \\
\end{tabular}
\caption{ Target translation distances of the surface from the reference model. Because a morphing cage approach is used for deformation, these minimum and maximum values are used as targets to inform the extent of the cage deformation, but are not precisely enforced.}
\label{table:suv_morph_params}
\end{table}


The geometry variants are divided into four categories based on vehicle size and AeroSUV configuration: full-scale estate, full-scale fastback, quarter-scale estate, and quarter-scale fastback, with each group containing 998 samples. Full-scale represents the geometry as described by the reference AeroSUV model while quarter-scale uniformly scales all dimensions by $0.25$. The quarter-scale dataset was used to validate the simulation setup by comparing to the experimental results acquired by \citep{zhang2021thesis} at FKFS using a $1/4$-scale wind tunnel model. The full-scale results leverage the same general approach, with adjustments based on best practices learned during AutoCFD \cite{economon2024autocfd} for full-scale vehicle simulations.

\begin{figure}[h!]
\centering
\begin{subfigure}{0.99\textwidth}
    \centering
    \includegraphics[width=0.99\linewidth]{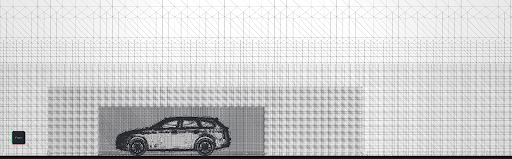}
\end{subfigure}%
\hfill
\begin{subfigure}{0.99\textwidth}
    \centering
    \includegraphics[width=0.99\linewidth]{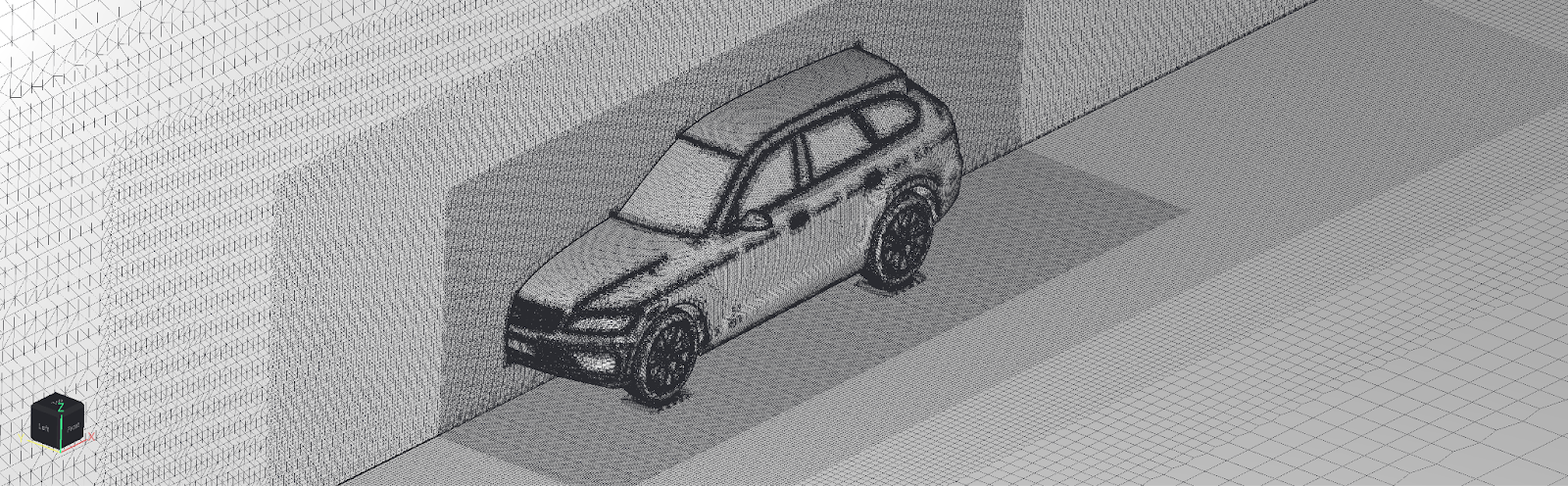}
\end{subfigure}
\caption{ Images depicting the mesh resolution refinement regions defined around the vehicle to capture the marge near-body and wake dynamics.
}
\label{fig:aerosuv_mesh}
\end{figure}

Computational meshes for the entire vehicle were generated in ANSA (each sample has a unique volume mesh), which generates hex-dominant and polyhedral meshes. A grid refinement study was performed for the validation configuration; refinement settings that produce $\approx45$M cells (see Figure~\ref{fig:aerosuv_mesh}) provided an acceptable balance of accuracy and computational efficiency.

Solutions are produced using the Luminary Cloud flow solver, a GPU-native finite-volume code which is second-order in both space and time~\cite{economon2024autocfd,krakos2025gpu}. The transient turbulence modeling approach leverages DDES with shear layer-adapted length scale and vortex titling measure (VTM) for grey area mitigation. All samples use the Spalart-Allmaras (SA) model for the Reynolds-averaged Navier Stokes (RANS) regions, and a hybrid centered/upwind convective scheme with proprietary blending allows for low dissipation in the large eddy simulation (LES) regions. An advanced shielding function avoids modeled stress depletion, preventing early prediction of separation.

A rolling floor and wheel rotation are modeled using translating and rotating surface boundary conditions, respectively. The rest of the vehicle surfaces are modeled as no-slip wall boundaries. A uniform inflow velocity of 30 m/s is used for the full-scale datasets, which are the only data used in the training described below.

\begin{figure}[h!]
\centering
\begin{subfigure}{0.99\textwidth}
    \centering
    \includegraphics[width=0.99\linewidth]{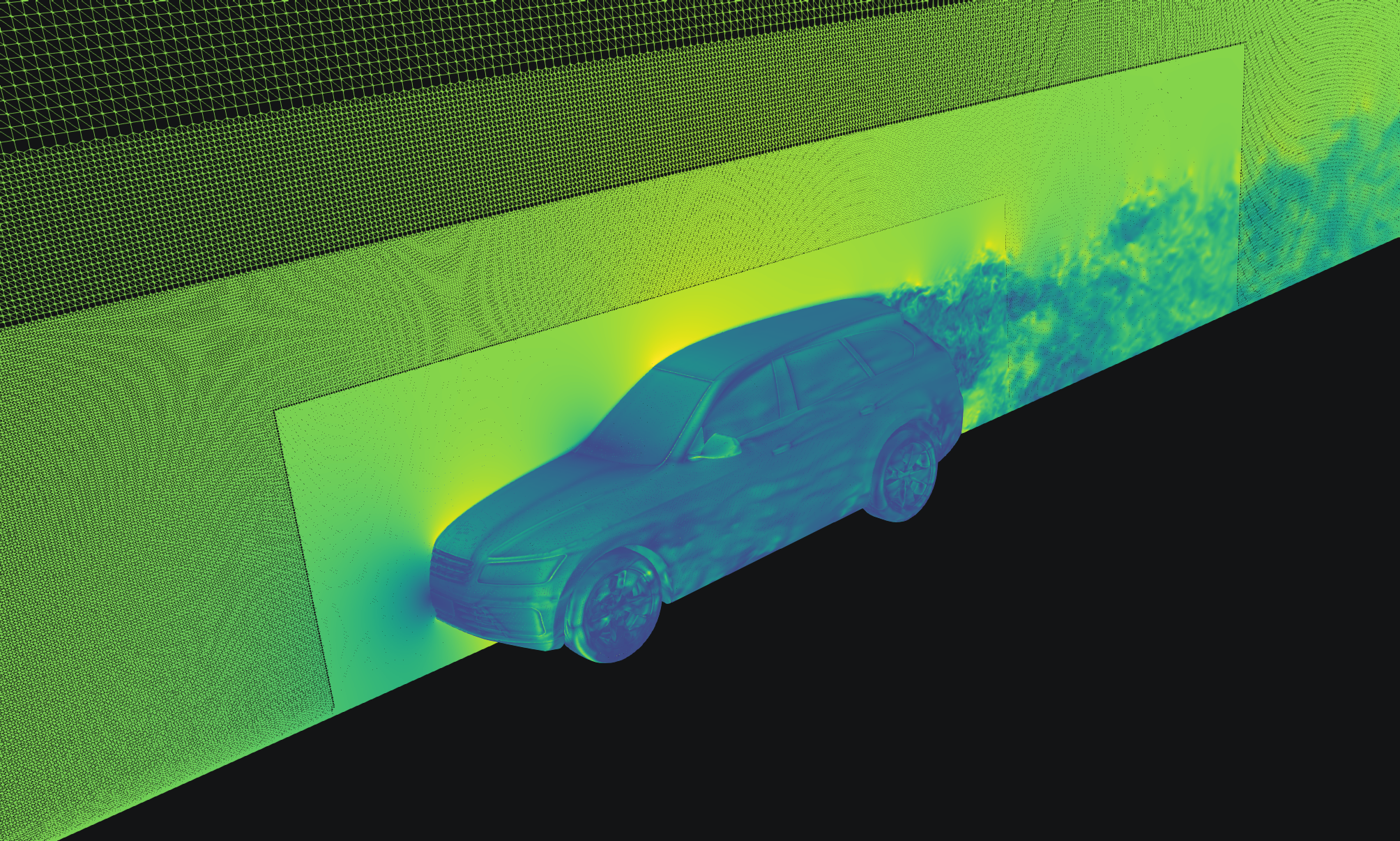}
\end{subfigure}
\caption{ Instantaneous velocity field (center plane) and wall shear stress contours (vehicle surface) for a full-scale Estate variant.
}
\label{fig:aerosuv_solution}
\end{figure}

\subsubsection{Training setup}
\label{sec:suv_setup}

We train AB-UPT for 250 epochs (400K updates) on the full-scale simulations of the SHIFT-SUV dataset. This amounts to roughly 1996 simulations which are randomly split with a 80/10/10 train/validation/test split. To avoid dataloading bottlenecks during training, we randomly subsample the volumetric data to 10\% of the original mesh cells. The training setting uses the LION optimizer~\cite{chen2023lion} with a peak learning rate of 5e-5 that is scheduled with a linear warmup the first 5\% of training followed by cosine decay to 1e-6. All models use 16K surface/volume tokens for their forward pass during training where AB-UPT does not use any query tokens during training (only for inference). Models are trained to predict all surface and volume variables at once where each variable is regressed via a mean absolute error loss function. We use a batch size of 1 and use the exponential moving average (update rate 1e-4) of the weights for evaluation. Training in float16 mixed precision takes roughly 13.5h on a single NVIDIA H100 GPU. Even though the training data contains two categories of cars (estate and fastback), we do not give this information explicitly to the model (e.g., via conditioning) as it is implicitly contained in the input geometry.

\subsubsection{Training data}

As described in Section~\ref{sec:dataset_suv} SHIFT-SUV contains 4 categories, full-scale estate, full-scale fastback, quarter-scale estate and quarter-scale fastback, with each category containing roughly 1000 simulations. We train on all full-scale simulations at once as we find that it is neither better nor faster to split it into fastback and estate car types. We tried training on all simulations at once and conditioning the model on the simulation scale, which ran into instability issues. We hypothesize that the different resolutions behave vastly different, leading to conflicting gradient directions which lead to training instabilities. Training models separately on all full-scale and all quater-scale simulations did not suffer from this. We choose the full-scale simulation resolution for this report to keep the results presentation concise. 
Table~\ref{table:suv_training_data} shows that training on the union of the estate and fastback car types at once performs better, suggesting synbiotic transfer of learned dynamics.

\begin{table}[h!]
\centering
\begin{tabular}{lccccccccc}
 & & \multicolumn{4}{c}{{Estate}} & \multicolumn{4}{c}{{Fastback}} \\ 
\cmidrule(rl){3-6} \cmidrule(rl){7-10} 
 & & \multicolumn{2}{c}{{Surface}} & \multicolumn{2}{c}{{Volume}} & \multicolumn{2}{c}{{Surface}} & \multicolumn{2}{c}{{Volume}} \\
\cmidrule(rl){3-4} \cmidrule(rl){5-6} \cmidrule(rl){7-8} \cmidrule(rl){9-10} 
Training data & Updates & $p_s$ & $\bm{\tau}_w$ & $p_v$ & $\bm{u}$ & $p_s$ & $\bm{\tau}_w$ & $p_v$ & $\bm{u}$  \\
\midrule
Estate & 200K & 8.60 & 3.78 & 3.27 & 2.55 & - & - & - & - \\
Fastback & 200K & - & - & - & - & 8.38 & 3.70 & 2.98 & 2.28 \\
Estate + fastback & 200K & 7.73 & 3.34 & 2.80 & 2.18 & 7.70 & 3.36 & 2.68 & 2.12 \\
Estate + fastback & 400K & \textbf{7.37} & \textbf{3.18} & \textbf{2.72} & \textbf{2.06} & \textbf{7.37} & \textbf{3.21} & \textbf{2.61} & \textbf{2.01}  \\
\end{tabular}
\caption{ Mean absolute errors of AB-UPT models trained on different data subsets  of SHIFT-SUV. Median performance over 5 seeds is reported. Training on both estate and fastback car types shows best performances. Errors are multiplied by 100 for $\bm{\tau}_w$ and 10 for $\bm{u}$.  }
\label{table:suv_training_data}
\end{table}

\subsubsection{Comparison against state-of-the-art baselines}

We compare AB-UPT against other transformer-based neural surrogate models in Table~\ref{table:suv_baselines}. AB-UPT performs best, obtaining good models in less than a day of training.

\begin{table}[h!]
\centering
\begin{tabular}{lcccccccccc}
 & & & \multicolumn{4}{c}{{Estate}} & \multicolumn{4}{c}{{Fastback}} \\ 
\cmidrule(rl){4-7} \cmidrule(rl){8-11} 
& & & \multicolumn{2}{c}{{Surface}} & \multicolumn{2}{c}{{Volume}} & \multicolumn{2}{c}{{Surface}} & \multicolumn{2}{c}{{Volume}} \\
\cmidrule(rl){4-5} \cmidrule(rl){6-7} \cmidrule(rl){8-9} \cmidrule(rl){10-11} 
 & GPU-hours & dim & $p_s$ & $\bm{\tau}_w$ & $p_v$ & $\bm{u}$ & $p_s$ & $\bm{\tau}_w$ & $p_v$ & $\bm{u}$  \\
\midrule
DoMINO & 35 & 32 & 42.85 & 21.25 & 682 & 15.26 & 42.33 & 21.47 & 687 & 15.09 \\
Transolver & 9.5 & 192 & 8.04 & 3.53 & 2.96 & 2.35 & 8.00 & 3.55 & 2.83 & 2.29 \\
Transformer & 16.5 & 192 & 7.70 & 3.35 & 2.78 & 2.17 & 7.70 & 3.37 & 2.66 & 2.12 \\
AB-UPT & 13.5 & 192 & \textbf{7.37} & \textbf{3.18} & \textbf{2.72} & \textbf{2.06} & \textbf{7.37} & \textbf{3.21} & \textbf{2.61} & \textbf{2.01}  \\
\midrule
AB-UPT & 22.3 & 384 & \textbf{6.47} & \textbf{2.85} & \textbf{2.56} & \textbf{1.89} & \textbf{6.50} & \textbf{2.91} & \textbf{2.49} & \textbf{1.88} \\
\end{tabular}
\caption{ Mean absolute errors of transformer-based models on SHIFT-SUV. Median performance over 5 seeds is reported. AB-UPT consistently performs best. Errors are multiplied by 100 for $\bm{\tau}_w$ and 10 for $\bm{u}$. For better performance, we also train a larger AB-UPT model with 1 seed, which we use for the analysis and visualizations of the subsequent sections. GPU-hours denote the time it took to train the model on a single NVIDIA H100 GPU. }
\label{table:suv_baselines}
\end{table}

\subsection{Relative errors}\label{label:suv_realtive_errors}
We present the relative L1 and L2 errors for the best model in Table~\ref{table:suv_relative_errors}, as they can provide a more interpretable accuracy measure due to the normalization by the norm of all targets in the dataset.
In particular, the relative L1 error is defined as $\mathrm{L}^{\mathrm{rel}}_1 = \frac{\sum_{i=1}^{n} |y_i - \hat{y}_i|}{\sum_{i=1}^{n} |y_i|}$, where $\hat{y}_i$ are predictions and $y_i$ are target field values; and the relative L2 error is analogously defined as the ratio of the L2 norms of all $n$ prediction errors and all targets, respectively. 
\begin{table}[h!]
\centering
\begin{tabular}{lcccccccc}
 & \multicolumn{4}{c}{{Estate}} & \multicolumn{4}{c}{{Fastback}} \\ 
\cmidrule(rl){2-5} \cmidrule(rl){6-9} 
 & \multicolumn{2}{c}{{Surface}} & \multicolumn{2}{c}{{Volume}} & \multicolumn{2}{c}{{Surface}} & \multicolumn{2}{c}{{Volume}} \\
\cmidrule(rl){2-3} \cmidrule(rl){4-5} \cmidrule(rl){6-7} \cmidrule(rl){8-9} 
  & $p_s$ & $\bm{\tau}_w$ & $p_v$ & $\bm{u}$ & $p_s$ & $\bm{\tau}_w$ & $p_v$ & $\bm{u}$  \\
\midrule
$\mathrm{L}^{\mathrm{rel}}_1$ & 0.0064\% & 4.95\% & 0.0025\% & 2.25\% & 0.0064\% & 5.03\% & 0.0024\% & 2.21\% \\
$\mathrm{L}^{\mathrm{rel}}_2$ & 0.0142\% & 5.97\% & 0.0054\% & 2.83\% & 0.0142\% & 6.16\% & 0.0054\% & 2.83\% \\
\end{tabular}
\caption{Relative L1 and L2 errors of different fields from the best model (AB-UPT).}
\label{table:suv_relative_errors}
\end{table}




\subsubsection{Aerodynamic forces}
\begin{figure}[h!]
\centering
\begin{subfigure}{0.24\textwidth}
\centering
\includegraphics[width=\linewidth]{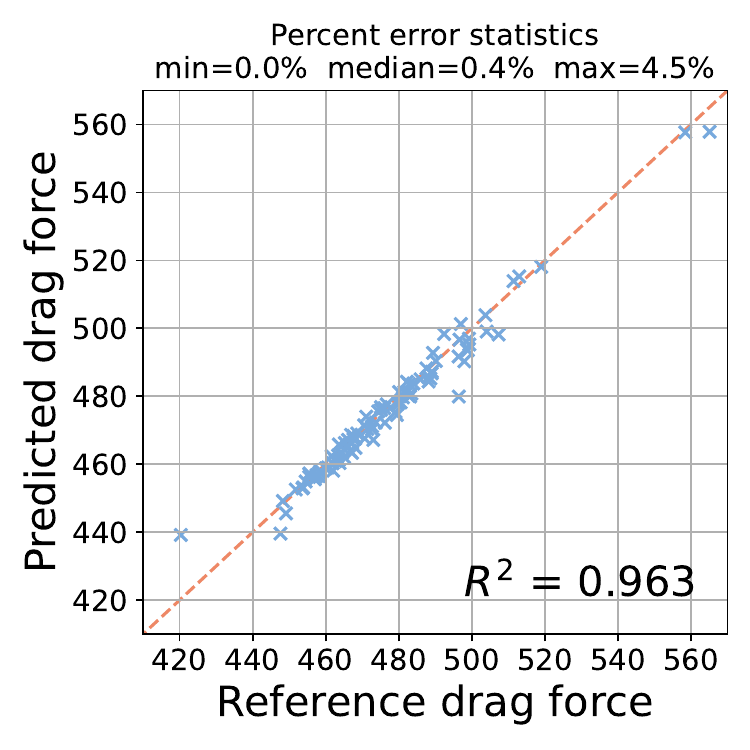}
\caption{Estate drag force}
\end{subfigure}%
\hfill
\begin{subfigure}{0.24\textwidth}
\centering
\includegraphics[width=\linewidth]{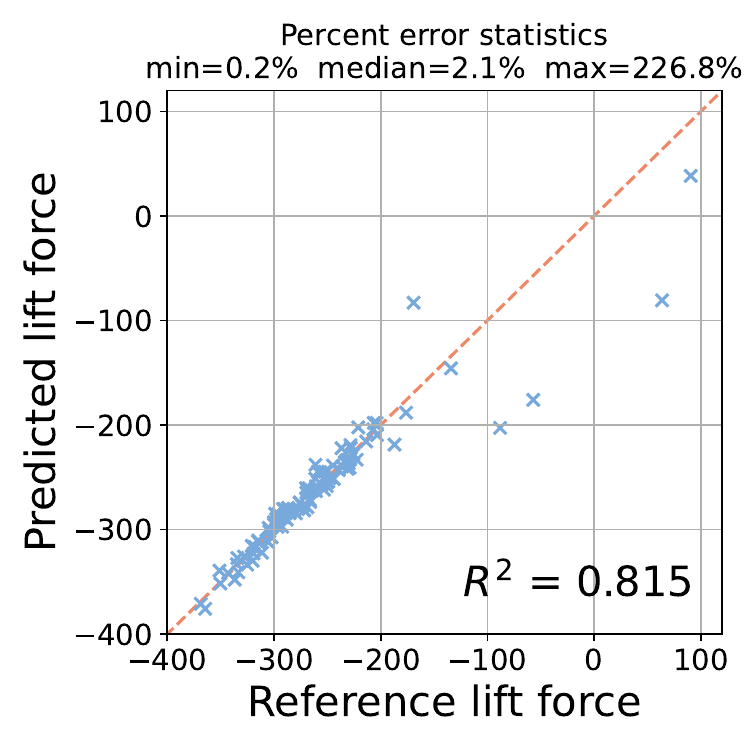}
\caption{Estate lift force}
\end{subfigure}%
\hfill
\begin{subfigure}{0.24\textwidth}
\centering
\includegraphics[width=\linewidth]{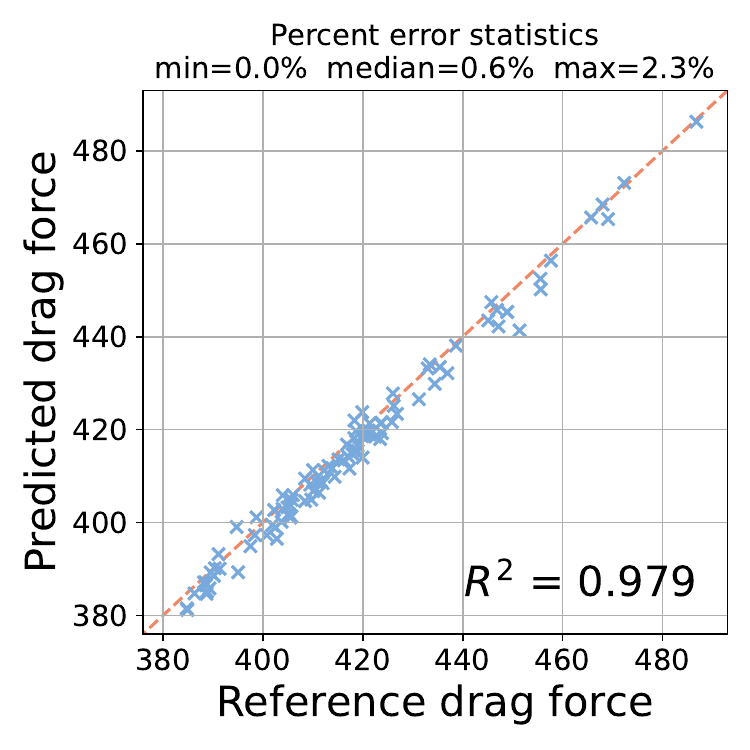}
\caption{Fastback drag force}
\end{subfigure}%
\hfill
\begin{subfigure}{0.24\textwidth}
\centering
\includegraphics[width=\linewidth]{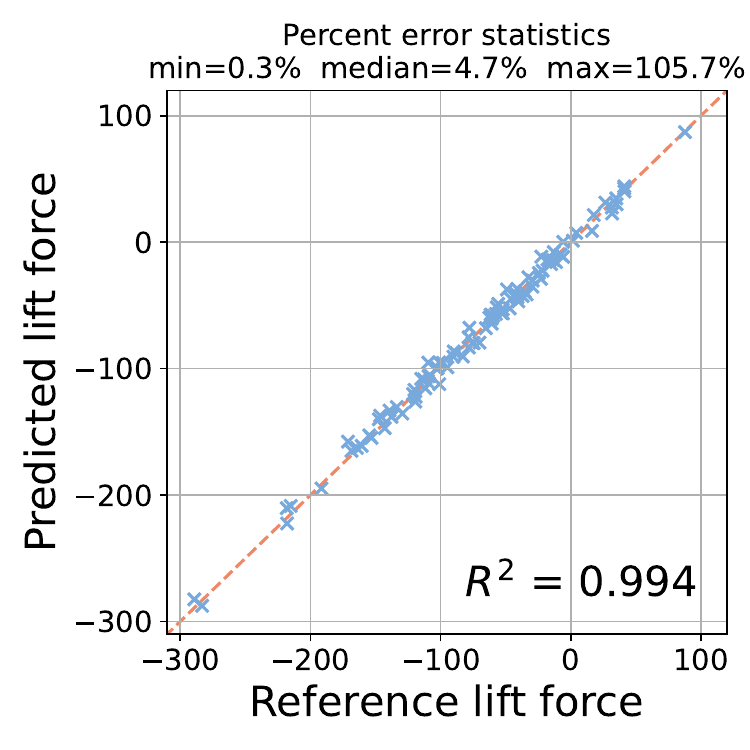}
\caption{Fastback lift force}
\end{subfigure}
\caption{ Aerodynamic drag and lift forces in Newtons. AB-UPT produces accurate predictions of these integrated quantities. The most difficult task is the lift force prediction for estate car types (b) where the bulk of simulations exhibit lift forces below -200~N (both train and test simulations), leading to larger errors than in the other tasks (a, c, d) where coverage of force magnitudes is more uniform. }
\label{fig:suv_coefficients_diagonal}
\end{figure}

We calculate the aerodynamic forces (Equation~\ref{eq:drag_and_lift_force}) by evaluating our best model (dim 384) at all positions of the surface mesh that was also used to run the numerical simulation. Figure~\ref{fig:suv_coefficients_diagonal} compares the predicted forces against the ground truth, i.e., the forces obtained from the surface variables of the numerical simulation.  AB-UPT is able to model aerodynamic forces with high accuracy where estate car types are more challenging than fastback car types. Notably, the lift forces of most estate car types is below -200~N where some samples of the testset also contain much larger lift forces (up to 100~N) where the model produces larger error. Note that we randomly assign samples to train/validation/test split, meaning that the number of estate training samples that the model has seen, which exhibit lift forces above -200~N, is quite small. Consequently, it is somewhat expected that such rare dynamics are not learned that well and performance could most likely be improved by including more estate geometries that result in lift forces greater than -200~N into the dataset.  


\subsubsection{Qualitative visualization}

\begin{figure}[h!]
\centering
\begin{subfigure}{0.3\textwidth}
    \centering
    \includegraphics[width=0.8\linewidth]{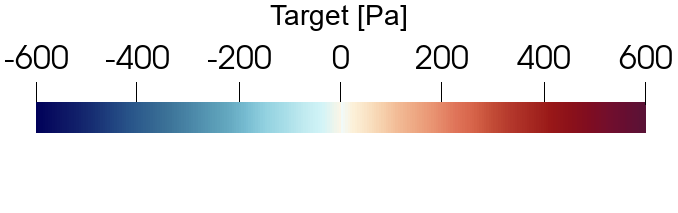}
\end{subfigure}%
\hfill
\begin{subfigure}{0.3\textwidth}
    \centering
    \includegraphics[width=0.8\linewidth]{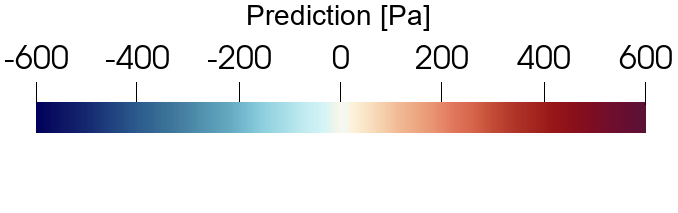}
\end{subfigure}%
\hfill
\begin{subfigure}{0.3\textwidth}
    \centering
    \includegraphics[width=0.8\linewidth]{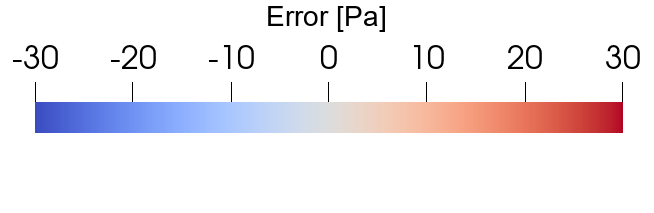}
\end{subfigure}
\begin{subfigure}{0.3\textwidth}
    \centering
    \includegraphics[width=\linewidth]{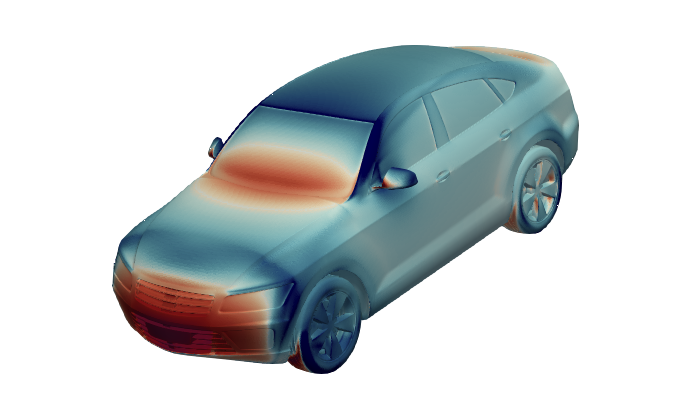}
\end{subfigure}%
\hfill
\begin{subfigure}{0.3\textwidth}
    \centering
    \includegraphics[width=\linewidth]{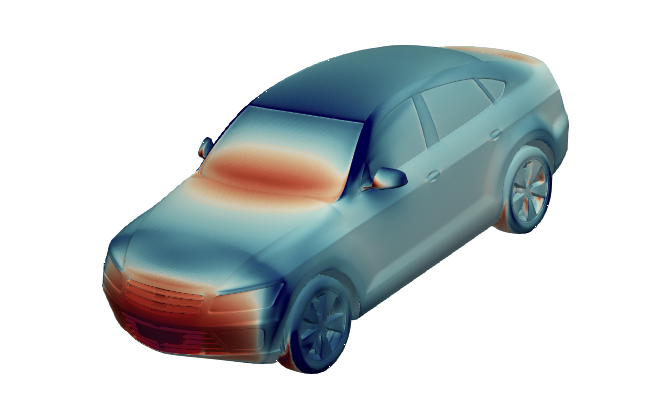}
\end{subfigure}%
\hfill
\begin{subfigure}{0.3\textwidth}
    \centering
    \includegraphics[width=\linewidth]{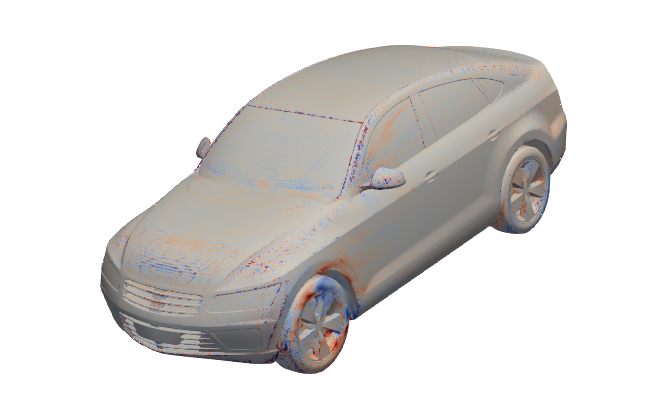}
\end{subfigure}
\begin{subfigure}{0.3\textwidth}
    \centering
    \includegraphics[width=\linewidth]{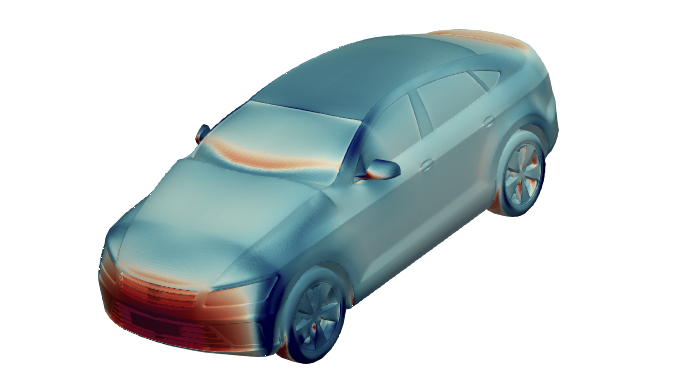}
\end{subfigure}%
\hfill
\begin{subfigure}{0.3\textwidth}
    \centering
    \includegraphics[width=\linewidth]{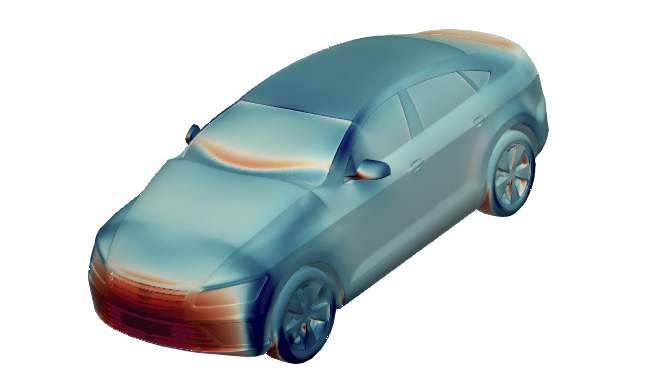}
\end{subfigure}%
\hfill
\begin{subfigure}{0.3\textwidth}
    \centering
    \includegraphics[width=\linewidth]{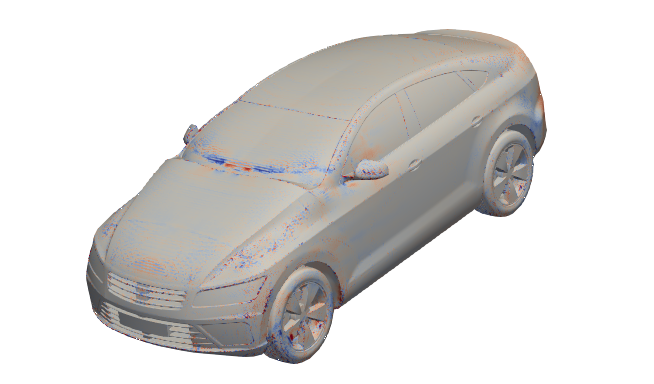}
\end{subfigure}
\begin{subfigure}{0.3\textwidth}
    \centering
    \includegraphics[width=\linewidth]{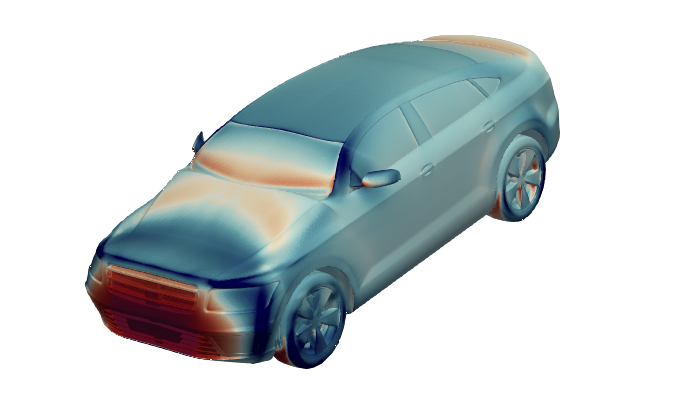}
\end{subfigure}%
\hfill
\begin{subfigure}{0.3\textwidth}
    \centering
    \includegraphics[width=\linewidth]{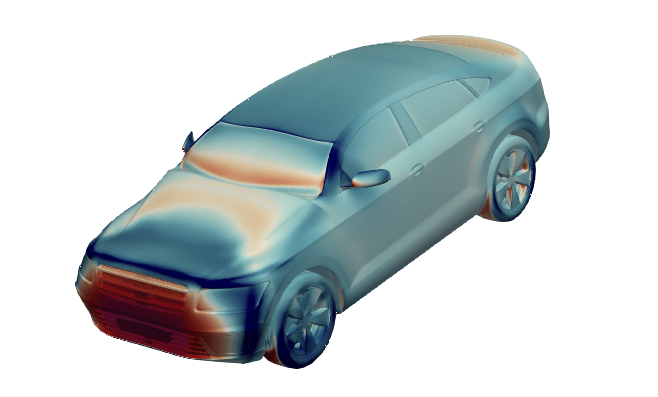}
\end{subfigure}%
\hfill
\begin{subfigure}{0.3\textwidth}
    \centering
    \includegraphics[width=\linewidth]{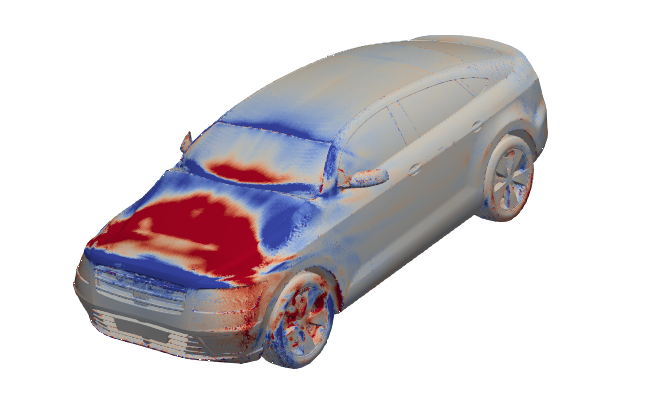}
\end{subfigure}
\caption{ Qualitative visualization of the surface pressure from the best (top), median (middle) and worst (bottom) prediction over all fastback simulations from the testset. Note that the colorbar of the error was chosen relatively strict to better show the errors made in the best and median case. Pressures are displayed as relative to the atmospheric pressure.}
\label{fig:suv_surface_visualization}
\end{figure}

\begin{figure}[h!]
\centering
\begin{subfigure}{0.3\textwidth}
    \centering
    \includegraphics[width=0.8\linewidth]{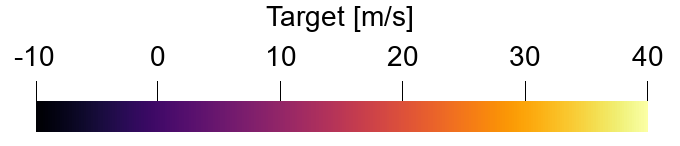}
\end{subfigure}%
\hfill
\begin{subfigure}{0.3\textwidth}
    \centering
    \includegraphics[width=0.8\linewidth]{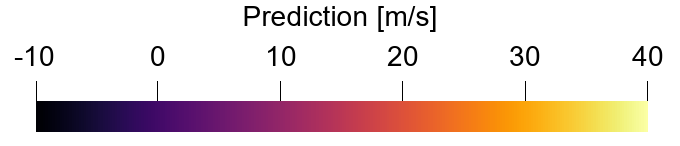}
\end{subfigure}%
\hfill
\begin{subfigure}{0.3\textwidth}
    \centering
    \includegraphics[width=0.8\linewidth]{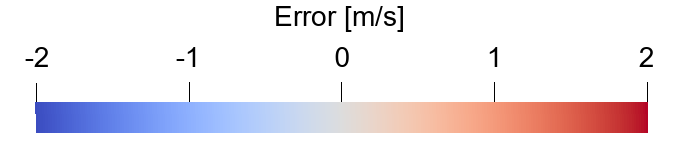}
\end{subfigure}
\begin{subfigure}{0.3\textwidth}
    \centering
    \includegraphics[width=\linewidth]{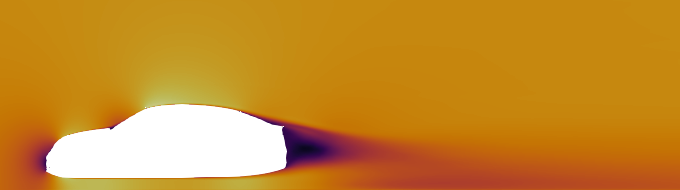}
\end{subfigure}%
\hfill
\begin{subfigure}{0.3\textwidth}
    \centering
    \includegraphics[width=\linewidth]{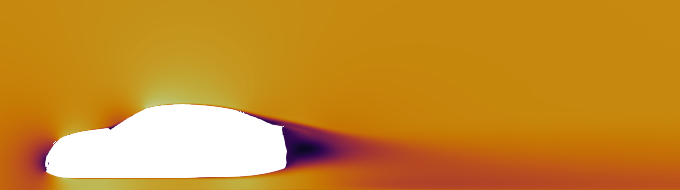}
\end{subfigure}%
\hfill
\begin{subfigure}{0.3\textwidth}
    \centering
    \includegraphics[width=\linewidth]{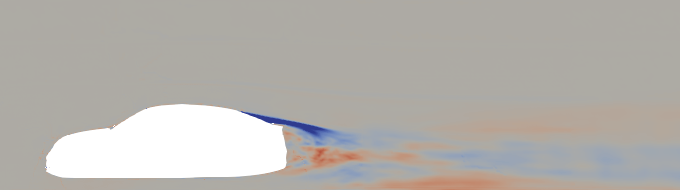}
\end{subfigure}
\begin{subfigure}{0.3\textwidth}
    \centering
    \includegraphics[width=\linewidth]{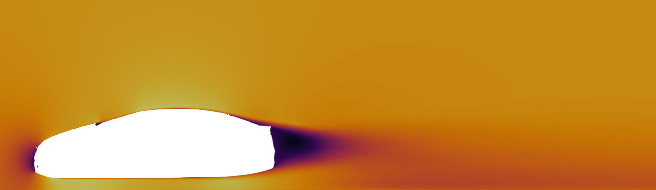}
\end{subfigure}%
\hfill
\begin{subfigure}{0.3\textwidth}
    \centering
    \includegraphics[width=\linewidth]{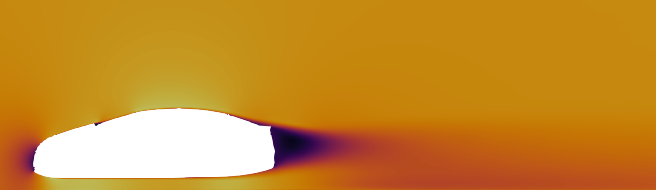}
\end{subfigure}%
\hfill
\begin{subfigure}{0.3\textwidth}
    \centering
    \includegraphics[width=\linewidth]{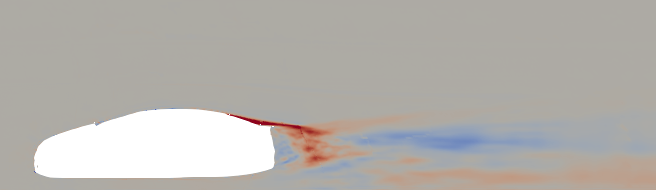}
\end{subfigure}
\begin{subfigure}{0.3\textwidth}
    \centering
    \includegraphics[width=\linewidth]{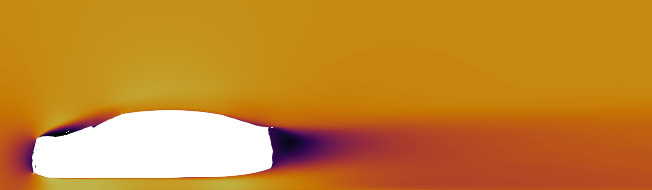}
\end{subfigure}%
\hfill
\begin{subfigure}{0.3\textwidth}
    \centering
    \includegraphics[width=\linewidth]{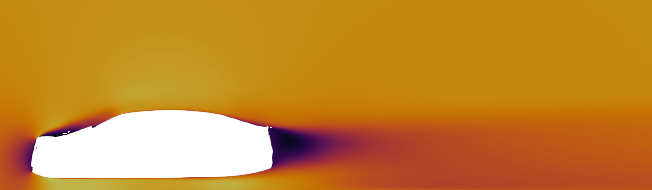}
\end{subfigure}%
\hfill
\begin{subfigure}{0.3\textwidth}
    \centering
    \includegraphics[width=\linewidth]{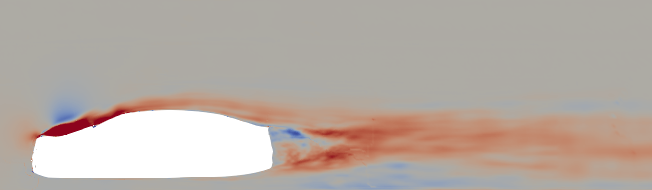}
\end{subfigure}
\caption{ Qualitative visualization of the x component of the velocity from the best (top), median (middle) and worst (bottom) prediction over all fastback simulations from the testset. 
}
\label{fig:suv_surface_visualization_velocity}
\end{figure}

Figure~\ref{fig:suv_surface_visualization} and Figure~\ref{fig:suv_surface_visualization_velocity} show qualitative visualizations of AB-UPT predictions with various error magnitudes. Target and prediction are visually indistinguishable, indicating good predictions where visualizing the error highlights mispredicted regions.

\subsubsection{Inference time analysis}

Table~\ref{table:suv_runtime} shows inference runtimes for various resolutions. Some use-cases (e.g., predicting only surface variables) can be calculated using lower runtimes as the flexible anchor-based design of AB-UPT necessitates 16K surface/volume tokens but any amount of query tokens is optional.

\begin{table}[h!]
\centering
\begin{tabular}{lccc}
 & \#Surface points & \#Volume points & Runtime [s] \\
\midrule
Full surface & 3M & 16K & 2.1 \\
Full volume & 16K & 45M & 31.5 \\
Full volume + surface & 3M & 45M & 33.6 \\
\end{tabular}
\caption{ Approximate inference runtime to create predictions for various number of output points. The number of surface and volume anchors is fixed to 16K. Anchor points are always required as the model uses cross-attention between surface and volume anchors in every other block. }
\label{table:suv_runtime}
\end{table}




\subsection{Luminary SHIFT-Wing}

\subsubsection{Dataset description}

SHIFT-Wing is an open-source database built around the NASA common research model (CRM)~\cite{nasa_crm} and is tailored for high-speed transonic commercial aircraft aerodynamics. The CRM has many different geometric configurations, ranging from high speed (cruise) to high lift (take-off/landing) with flaps and slats deployed, optional nacelle and pylon, optional horizontal stabilizers, and even vertical tail. These configurations have been thoroughly experimentally investigated and has inspired a history of CFD simulations through the AIAA Drag Prediction and High-Lift workshops. SHIFT-Wing is provided with the CC-BY-NC license and is available to download at no cost through HuggingFace~\cite{shiftwing2025}.

\begin{figure}[h!]
\centering
\begin{subfigure}{0.32\textwidth}
\centering
\includegraphics[width=0.99\linewidth]{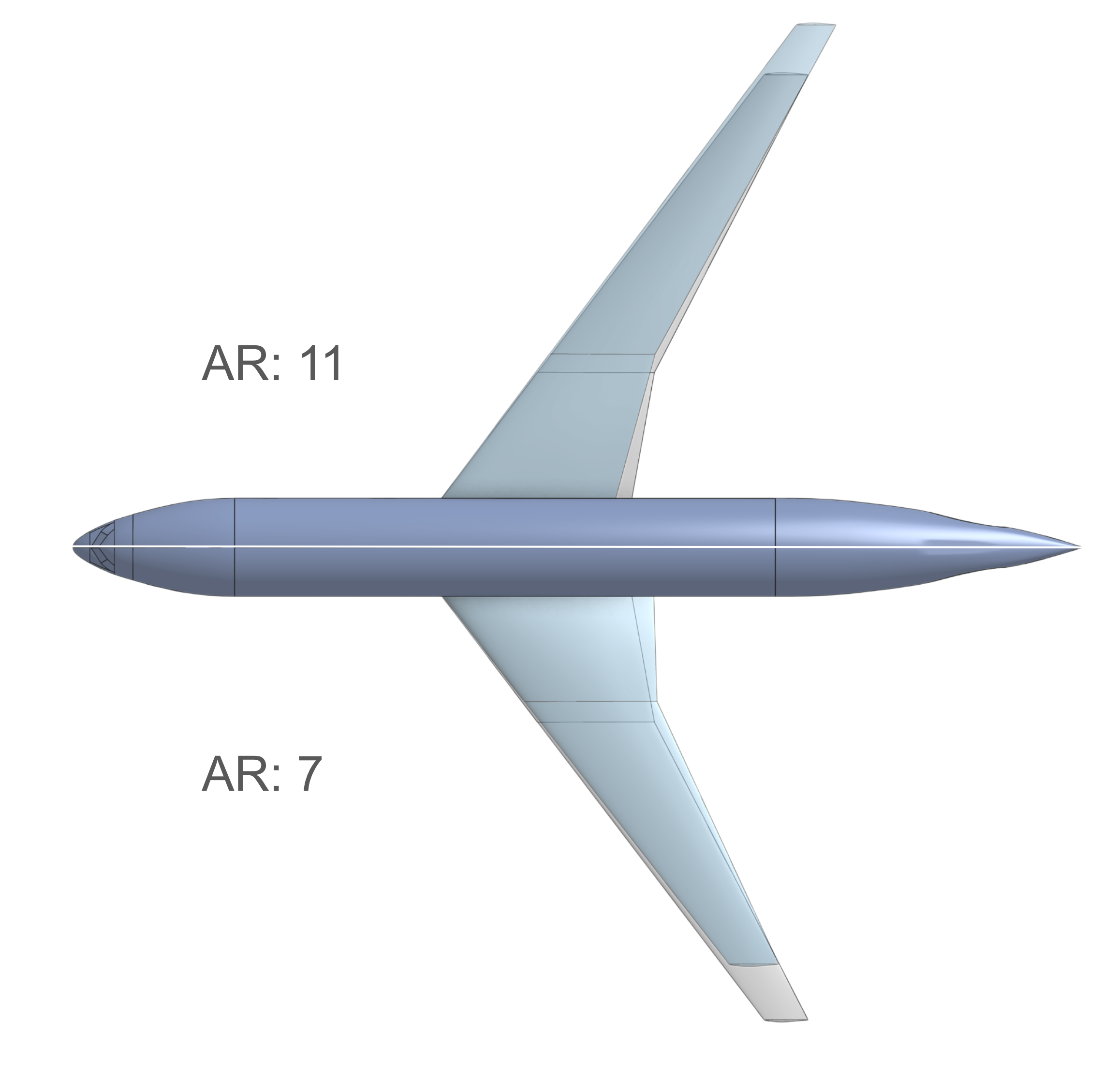}
\caption{Aspect ratio}
\end{subfigure}
\begin{subfigure}{0.32\textwidth}
\centering
\includegraphics[width=0.99\linewidth]{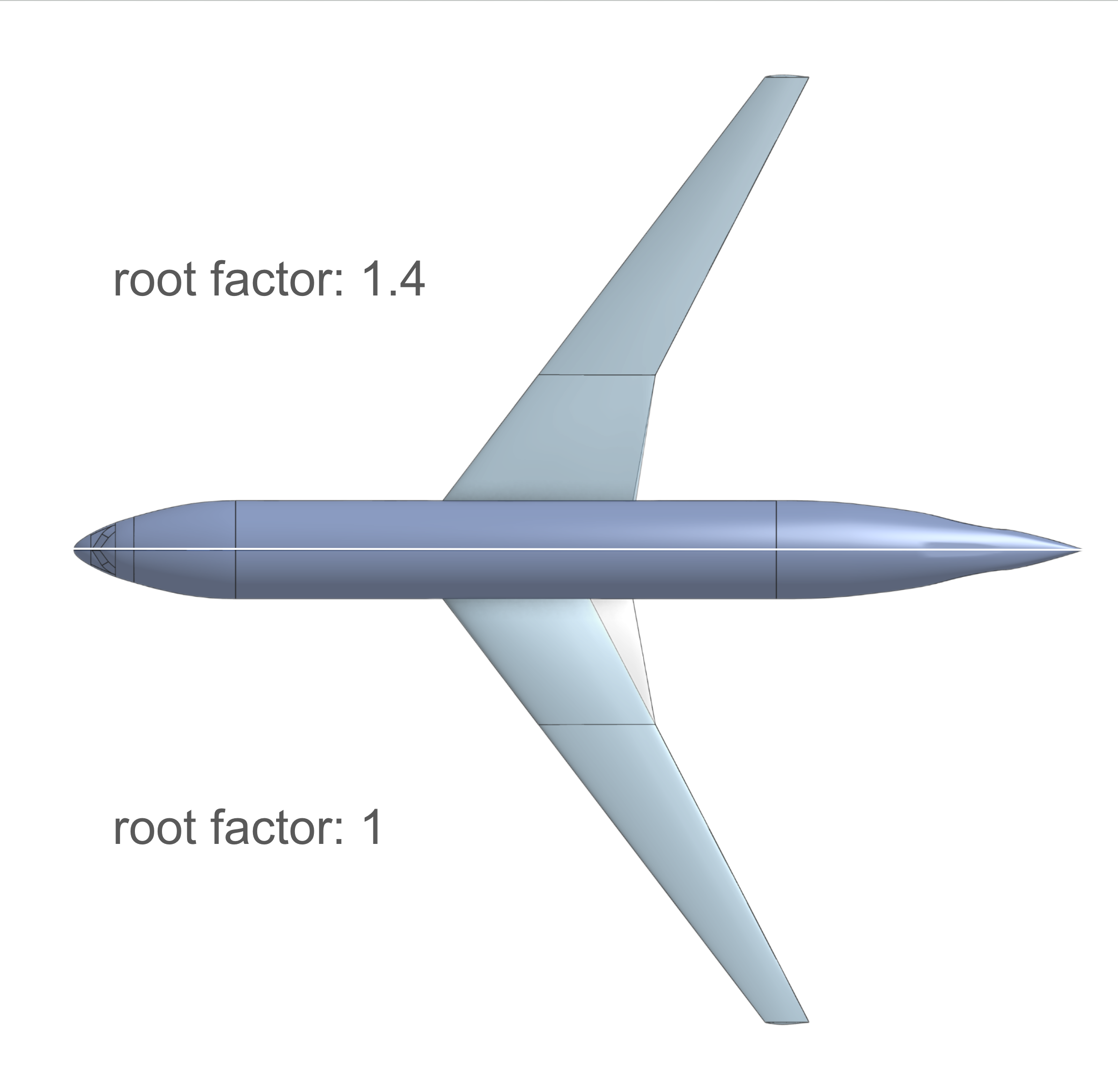}
\caption{$F_{C_{r}}$}
\end{subfigure}
\begin{subfigure}{0.32\textwidth}
\centering
\includegraphics[width=0.99\linewidth]{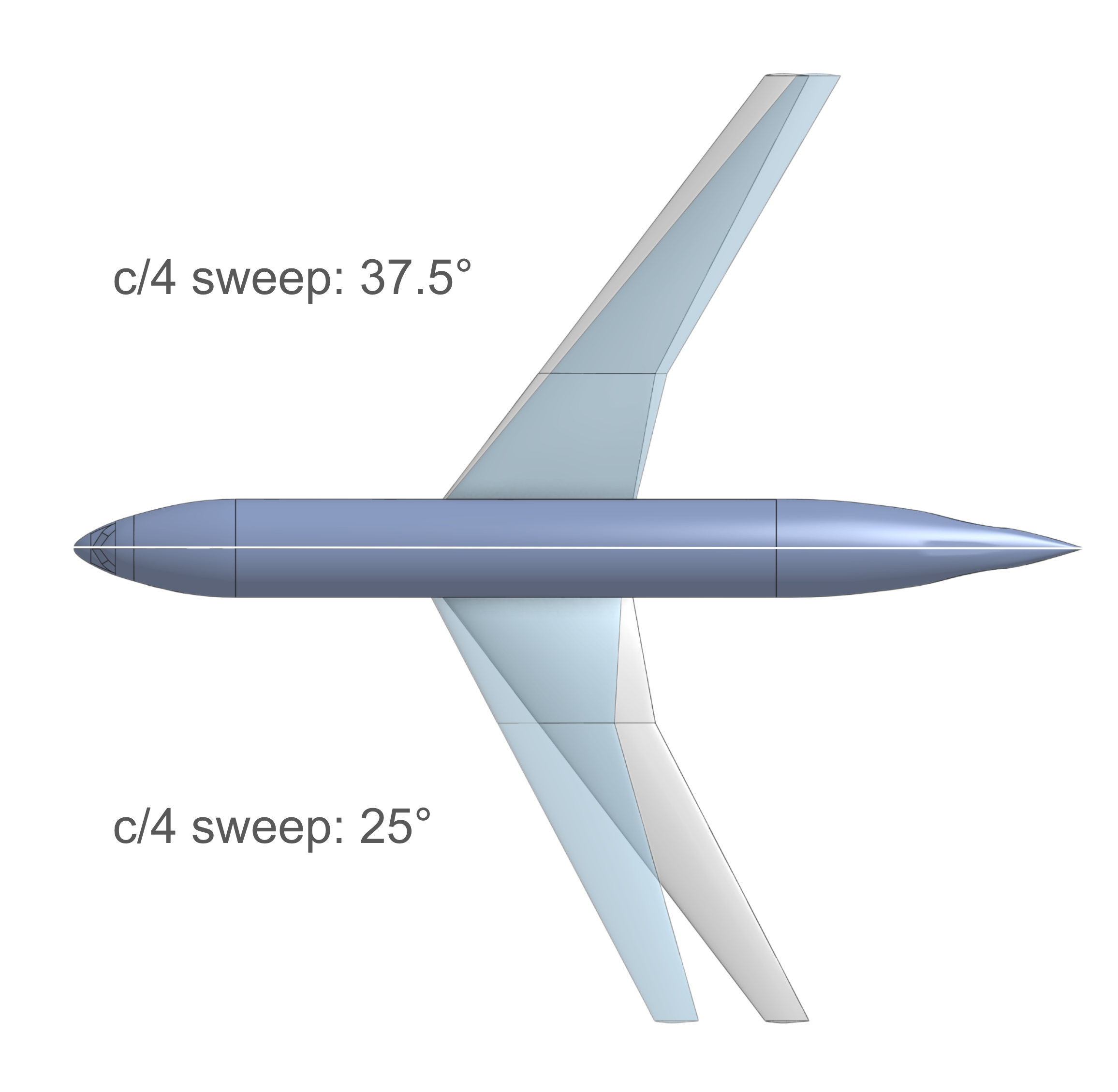}
\caption{$\beta_{\frac{1}{4}}$}
\end{subfigure}
\caption{Comparing realizations of the wing by modifying one parameter at a time. For each image the minimum (bottom half) and maximum (top half) range model is represented overlaid on the reference geometry (gray).}
\label{fig:wing_geo_variants_wing}
\end{figure}

\begin{figure}[h!]
\centering
\begin{subfigure}{0.8\textwidth}
\centering
\includegraphics[width=0.99\linewidth]{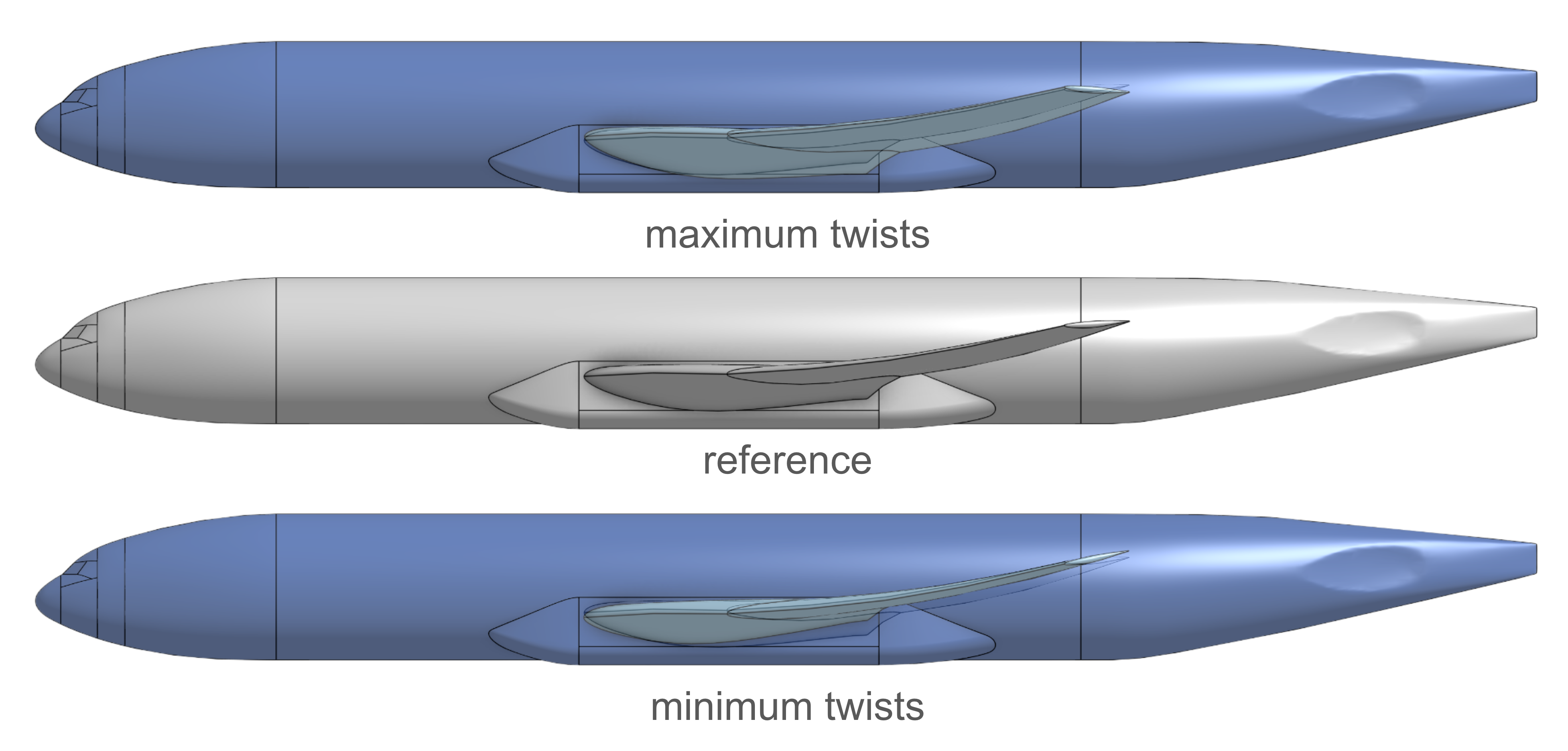}
\caption{twists}
\end{subfigure}
\caption{Comparison of the reference model (middle) with minimum range values for the twist parameters (bottom) and maximum range values (top). For both top and bottom images a 30\% opacity reference model is overlaid.}
\label{fig:wing_geo_variants_twist}
\end{figure}

The focus of the current dataset is the high-speed cruise configuration with only fuselage and wing, focusing primarily on planform design. A parametric model of the CRM was constructed in the geometry modeling software OnShape by importing the reference model, deconstructing it, and reconstructing with parameters introduced via OnShape's Variable Studio feature. This involved:
\begin{itemize}
    \item Splitting the wing and fuselage into separate parameterized models,
    \item Intersecting the reference wing at six span-wise locations and storing the resulting airfoil profiles,
    \item Exposing Variable Studio parameters to manipulate the translation and rotation (twist) of these profiles,
    \item Re-lofting the wing and reconstructing the wing-tip geometry,
    \item Parameterizing the length and radius of the fuselage,
    \item Parameterizing the size of the wing-body fairing based on the local chord of the new wing,
    \item Boolean operation to combine the wing and fuselage parts.
\end{itemize}
While there are many ``micro'' parameters that are used to construct the new model, the dataset is constructed by manipulating high-level ``macro'' parameters, described in Table~\ref{table:wing_params}. These parameters are commonly used for preliminary planform design, and the micro-parameters are computed from these. In the current dataset seven parameters describe the wing and fuselage, resulting in a diverse array of geometries as demonstrated in Figures~\ref{fig:wing_geo_variants_wing} and~\ref{fig:wing_geo_variants_twist}. 

\begin{figure}[h!]
\centering
\begin{subfigure}{0.8\textwidth}
\centering
\includegraphics[width=0.99\linewidth]{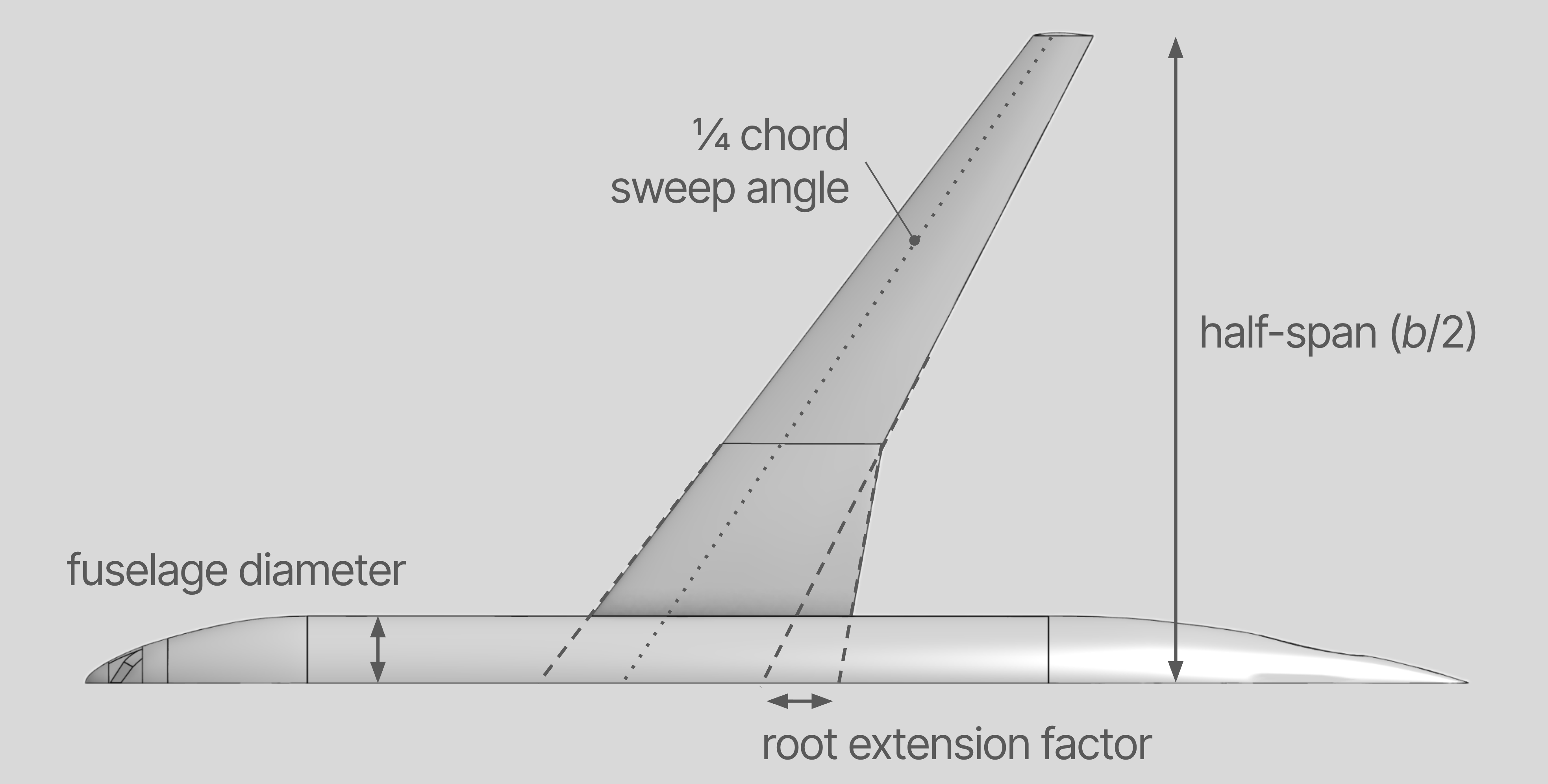}
\end{subfigure}
\caption{Diagram describing the macro parameters for the wing planform used to define the wing parameterization.}
\label{fig:wing_geo_diagram}
\end{figure}

\begin{figure}[h!]
\centering
\begin{subfigure}{0.9\textwidth}
\centering
\includegraphics[width=0.99\linewidth]{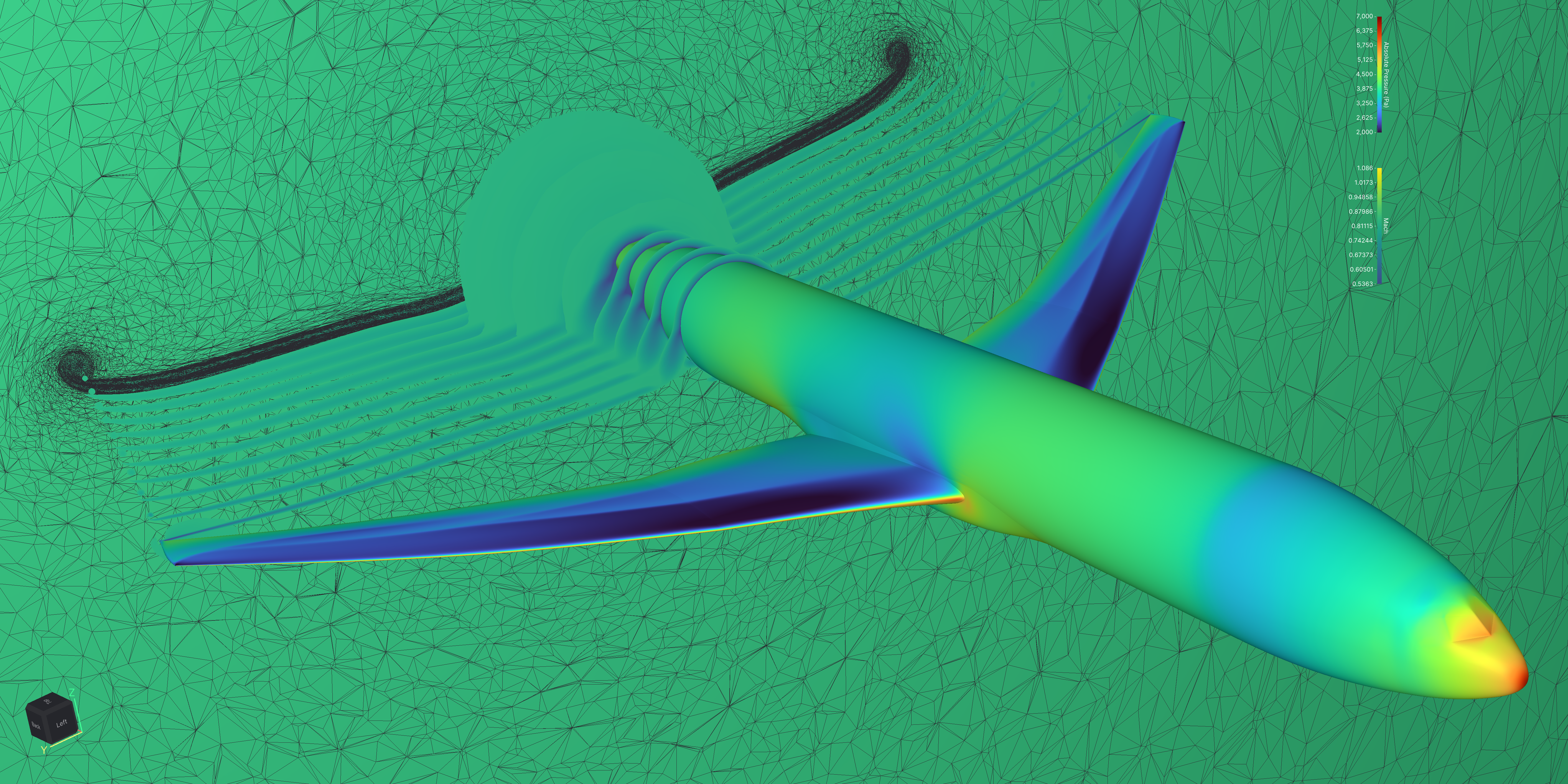}
\end{subfigure}
\caption{ Single solution depicting pressure contours on the surface of the airplane and a depiction of the solution-adapted mesh on the $x$-aligned plane downstream (behind) the vehicle. }
\label{fig:wing_geo_solution}
\end{figure}

The dataset is generated in batches at fixed flow speed (Mach number), where the remaining parameters (geometry and angle of attack) are sampled via Latin hypercube sampling. This is done intentionally to facilitate assessment of AI/ML approaches on fundamentally different flow physics. The lower speeds do not induce shocks, sharp discontinuities in the flow field, while the higher Mach number conditions produce complex 3-dimensional shock patterns (see contours in Figure~\ref{fig:wing_geo_solution}).

\begin{table}[h!]
\centering
\begin{tabular}{llccc}
design space parameter & symbol & Min & Ref & Max \\
\midrule
Aspect Ratio & AR  & 7.5 & 9 & 11 \\
1/4 chord sweep angle & $\beta_{\frac{1}{4}}$ & 25$^\circ$ & 35$^\circ$ & 37.5$^\circ$ \\
Root-chord extension factor & $F_{C_{r}}$ & 1.0 & 1.373 & 1.4 \\
Fuselage diameter & $D_f$ & 240 in & 240 in & 258 in \\
Root twist & $t_{r}$ & 3$^\circ$ & 6.717$^\circ$ & 9$^\circ$ \\
Delta twist root-to-break & $\Delta t_{r,b}$ & -7$^\circ$ & -5.953$^\circ$ & -3$^\circ$ \\
Delta twist break-to-tip & $\Delta t_{b,t}$ & -7.5$^\circ$ & -4.513$^\circ$ & -1.5$^\circ$ \\
\midrule
angle of attack & $\alpha$ & 0$^\circ$ &  & 4$^\circ$ \\
Mach number & $\textrm{Ma}$ & 0.5 &  & 0.85 \\
\end{tabular}
\caption{SHIFT-Wing design space parameters and their ranges relative to the reference geometry values, see Figure~\ref{fig:wing_geo_diagram}.}
\label{table:wing_params}
\end{table}

The same Luminary flow solver described in Section~\ref{sec:dataset_suv} is used for these simulations. Turbulence is characterized via steady RANS with the Spalart-Allmaras turbulence model~\cite{krakos2025gpu}. All vehicle surfaces are modeled as no-slip walls, with the angle-of-attack imposed via far field boundary conditions.

An important difference relative to the SHIFT-SUV dataset is the meshing strategy. Luminary Mesh Adaptation, a proprietary solution-adaptive meshing technology, is leveraged to automatically capture sharp flow features, as demonstrated in in Figure~\ref{fig:wing_geo_solution}. This solution procedure generates a sequence of meshes and solutions, automatically adapting each subsequent mesh (adjusting local mesh density and anisotropy) to improve accuracy. Sharp flow features, which are incredibly hard or computationally expensive to capture with manual meshing approaches, are precisely characterized for every sample of this dataset without user intervention.

\subsubsection{Training setup}

We train two AB-UPT on the SHIFT-Wing dataset where separate models are trained for Mach 0.5 and Mach 0.85 as we did not observe improved performance when training on both mach numbers at once (see Table~\ref{table:wing_training_data}).
This amounts to 1138 simulations for Mach 0.5 and 560 simulations for Mach 0.85.
We employ the same preprocessing and training protocol as for SHIFT-SUV. Namely, we randomly split the dataset into 80/10/10 train/validation/test samples, randomly subsample the volumetric data to 10\% for improved dataloading speed, train with the LION optimizer~\cite{chen2023lion}, use a peak learning rate of 5e-5 with a learning rate warmup of 5\% followed by cosine decay to 1e-6, train for 400K updates, use a batch size of 1 with a model weight exponential moving average update rate of 1e-4 and train the model to predict all surface/volume variables at once with a mean absolute error.
Different from the SHIFT-SUV protocol, we train in float32 precision as we find it performs much better and use only 8K surface/volume anchors to compensate for the increased training time required for float32 training. Training takes roughly 25h on a single NVIDIA H100 GPU. As the training data varies the angle of attack $\alpha$ per simulation, we integrate $\alpha$ through Diffusion Transformer (DiT) style conditioning~\cite{peebles22dit} of all blocks in the surface and volume branches.

\subsubsection{Training data}

SHIFT-Wing contains 2 categories, Mach 0.5 and Mach 0.85 where we train separate models per Mach number. We tried training on both categories at once and conditioning the model on the mach number, which did not improve performance. We hypothesize that these categories exhibit significantly different dynamics, prohibiting symbiotic transfer between categories. Table~\ref{table:wing_training_data} shows that separate models for different mach numbers outperform a model trained on the union.


\begin{table}[h!]
\centering
\begin{tabular}{cccccccccc}
& & \multicolumn{4}{c}{{Mach 0.5}} & \multicolumn{4}{c}{{Mach 0.85}} \\ 
\cmidrule(rl){3-6} \cmidrule(rl){7-10} 
& & \multicolumn{2}{c}{{Surface}} & \multicolumn{2}{c}{{Volume}} & \multicolumn{2}{c}{{Surface}} & \multicolumn{2}{c}{{Volume}} \\
\cmidrule(rl){3-4} \cmidrule(rl){5-6} \cmidrule(rl){7-8} \cmidrule(rl){9-10} 
Mach & Updates & $p_s$ & $\bm{\tau}_w$ & $p_v$ & $\bm{u}$ & $p_s$ & $\bm{\tau}_w$ & $p_v$ & $\bm{u}$  \\
\midrule
0.5 & 200K & 5.60 & 0.550 & 6.84 & 2.38 & - & - & - & - \\
0.5 & 400K & \textbf{5.24} & \textbf{0.549} & \textbf{6.46} & \textbf{2.30} & - & - & - & - \\
0.85 & 200K & - & - & - & - & 18.64 & 1.462 & 29.37 & 4.35 \\
0.85 & 400K & - & - & - & - & \textbf{17.72} & \textbf{1.459} & \textbf{28.12} & \textbf{4.23} \\
0.5 \& 0.85 & 400K & 5.70 & 0.551 & 6.79 & 2.32 & 18.89 & 1.461 & 30.13 & 4.25 \\
\end{tabular}
\caption{ Mean absolute errors of AB-UPT models trained on different data subsets of SHIFT-Wing. Median performance over 5 seeds is reported. Best results are marked bold and second best are underlined. Training separate models per Mach number shows best performance. }
\label{table:wing_training_data}
\end{table}

\subsubsection{Comparison against state-of-the-art baselines}

We compare AB-UPT against other transformer-based neural surrogate models in Table~\ref{table:wing_baselines}. AB-UPT obtains good models in half a day where training can be prolonged for better performance.

\begin{table}[h!]
\centering
\begin{tabular}{lcccccccccc}
& & & \multicolumn{4}{c}{{Mach 0.5}} & \multicolumn{4}{c}{{Mach 0.85}} \\ 
\cmidrule(rl){4-7} \cmidrule(rl){8-11} 
& & & \multicolumn{2}{c}{{Surface}} & \multicolumn{2}{c}{{Volume}} & \multicolumn{2}{c}{{Surface}} & \multicolumn{2}{c}{{Volume}} \\
\cmidrule(rl){4-5} \cmidrule(rl){6-7} \cmidrule(rl){8-9} \cmidrule(rl){10-11} 
& GPU-hours & Updates & $p_s$ & $\bm{\tau}_w$ & $p_v$ & $\bm{u}$ & $p_s$ & $\bm{\tau}_w$ & $p_v$ & $\bm{u}$  \\
\midrule
DoMINO & 24 & 50K & 207 & 0.547 & 260 & 2.98 & 743 & 1.62 & 745 & 9.49 \\
Transolver & 4.7 & 200K & 8.24 & 0.559 & 9.51 & 3.04 & 30.7 & 1.486 & 46.1 & 6.77 \\
Transformer & 16.3 & 200K & 5.78 & 0.551 & 6.87 & 2.37 & 19.3 & 1.464 & 29.8 & 4.40 \\
AB-UPT & 12.5 & 200K & \textbf{5.60} & \textbf{0.550} & \textbf{6.84} & \textbf{2.38} & \textbf{18.6} & \textbf{1.461} & \textbf{29.4} & \textbf{4.35} \\
\hline
AB-UPT & 25.0 & 400K & \textbf{5.24} & \textbf{0.549} & \textbf{6.46} & \textbf{2.30} & \textbf{17.7} & \textbf{1.459} & \textbf{28.1} & \textbf{4.23} \\
\end{tabular}
\caption{ Mean absolute errors of AB-UPT models trained on SHIFT-Wing. Median performance over 5 seeds is reported. For better performance, we also train a AB-UPT model for longer, which we use for the analysis and visualizations of the subsequent sections. GPU-hours denote the time it took to train the model on a single NVIDIA H100 GPU. }
\label{table:wing_baselines}
\end{table}

\subsection{Relative errors}
Analogous to Section~\ref{label:suv_realtive_errors}, we present the relative L1 and L2 errors for the best model in Table~\ref{table:wing_relative_errors} in order to provide a more interpretable measure due to the normalization by the norm of the targets.
\begin{table}[h!]
\centering
\begin{tabular}{lcccccccc}
 & \multicolumn{4}{c}{{Mach 0.5}} & \multicolumn{4}{c}{{Mach 0.85}} \\ 
\cmidrule(rl){2-5} \cmidrule(rl){6-9} 
& \multicolumn{2}{c}{{Surface}} & \multicolumn{2}{c}{{Volume}} & \multicolumn{2}{c}{{Surface}} & \multicolumn{2}{c}{{Volume}} \\
\cmidrule(rl){2-3} \cmidrule(rl){4-5} \cmidrule(rl){6-7} \cmidrule(rl){8-9} 
 & $p_s$ & $\bm{\tau}_w$ & $p_v$ & $\bm{u}$ & $p_s$ & $\bm{\tau}_w$ & $p_v$ & $u$  \\
\midrule
$\mathrm{L}^{\mathrm{rel}}_1$ & 0.022\% & 12.5\% & 0.027\% & 9.56\% & 0.079\% & 13.3\% & 0.125\% & 9.51\%  \\
$\mathrm{L}^{\mathrm{rel}}_2$ & 0.056\% & 18.4\% & 0.072\% & 12.45\% & 0.221\% & 18.8\% & 0.651\% & 12.60\% 
\end{tabular}
\caption{Relative L1 and L2 errors of different fields from the best model (AB-UPT).}
\label{table:wing_relative_errors}
\end{table}

%
%

\subsubsection{Aerodynamic forces}

\begin{figure}[h!]
\centering
\begin{subfigure}{0.24\textwidth}
\centering
\includegraphics[width=\linewidth]{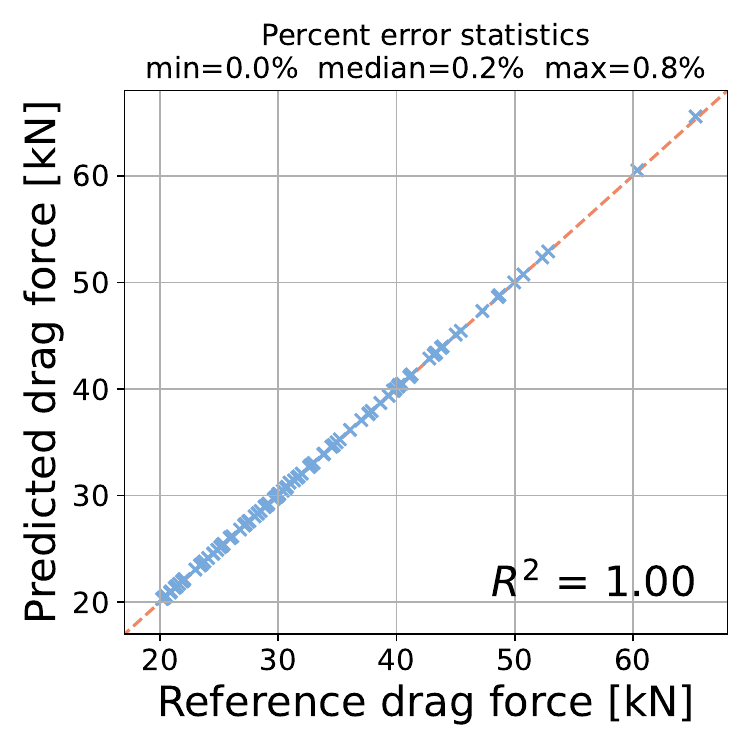}
\caption{Mach 0.5 drag force}
\end{subfigure}%
\hfill
\begin{subfigure}{0.24\textwidth}
\centering
\includegraphics[width=\linewidth]{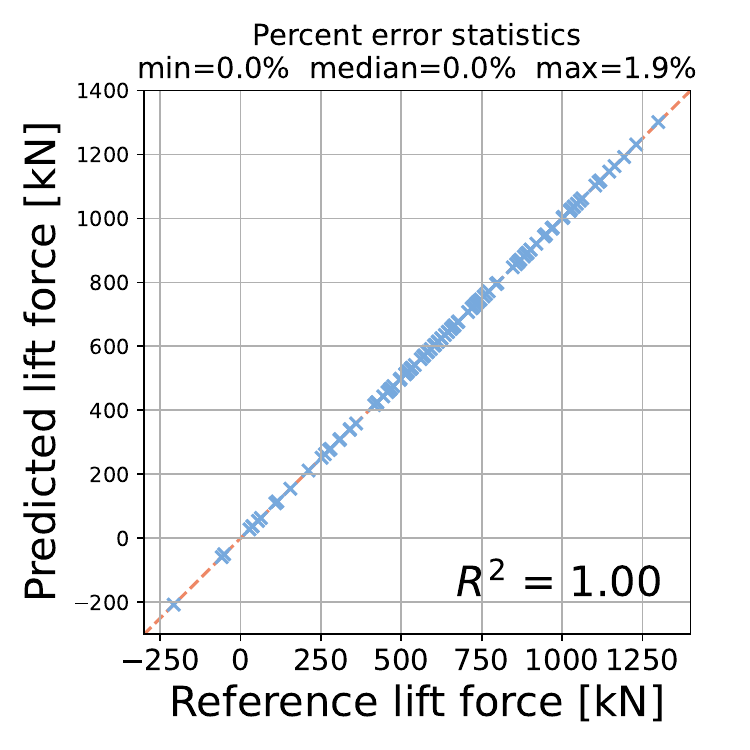}
\caption{Mach 0.5 lift force}
\end{subfigure}%
\hfill
\begin{subfigure}{0.24\textwidth}
\centering
\includegraphics[width=\linewidth]{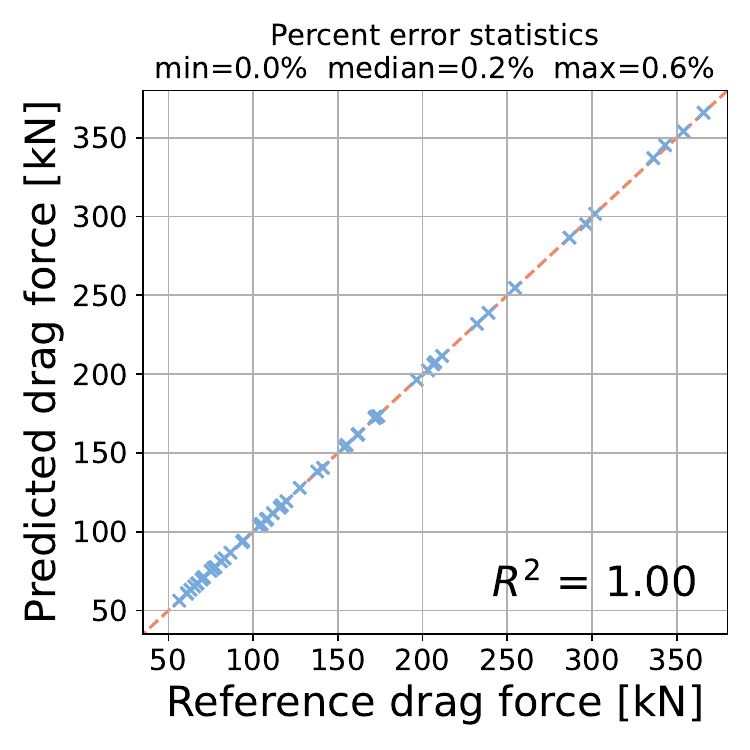}
\caption{Mach 0.85 drag force}
\end{subfigure}%
\hfill
\begin{subfigure}{0.24\textwidth}
\centering
\includegraphics[width=\linewidth]{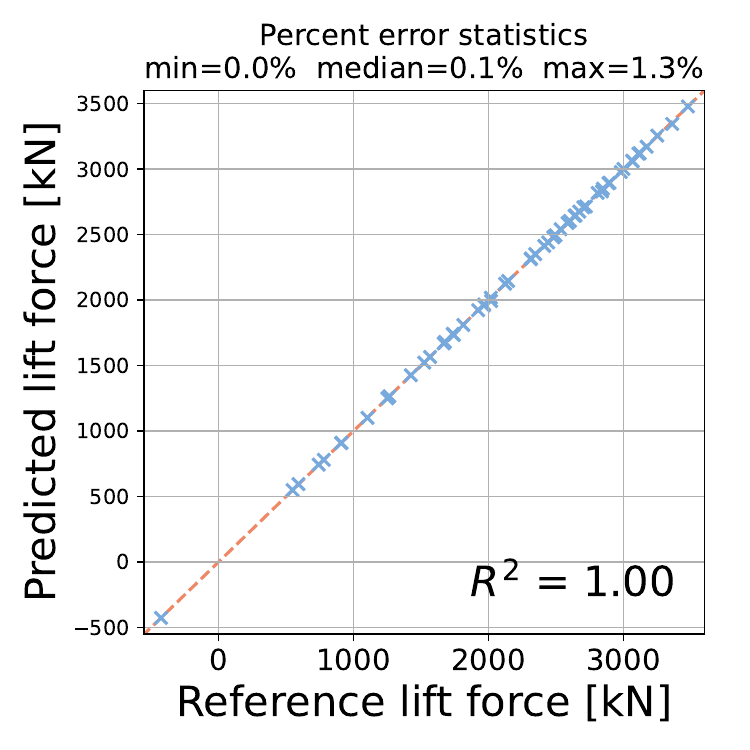}
\caption{Mach 0.85 lift force}
\label{fig:mach085_lift_force}
\end{subfigure}
\caption{ Aerodynamic drag and lift forces. AB-UPT can model these integrated quantities almost perfectly, obtaining an $R^2$ score of 1.0 with a worst-case error of at most 2\%.}
\label{fig:wing_coefficients_diagonal}
\end{figure}

We calculate the aerodynamic forces (Equation~\ref{eq:drag_and_lift_force}) by evaluating our best model at all positions of the surface mesh that was also used to run the numerical simulation. Figure~\ref{fig:wing_coefficients_diagonal} compares the predicted forces against the ground truth, i.e., the forces obtained from the surface variables of the numerical simulation. AB-UPT is able to model aerodynamic forces on the testset almost perfectly.


\begin{figure}[h!]
\centering
\begin{subfigure}{0.19\textwidth}
\centering
\includegraphics[width=\linewidth]{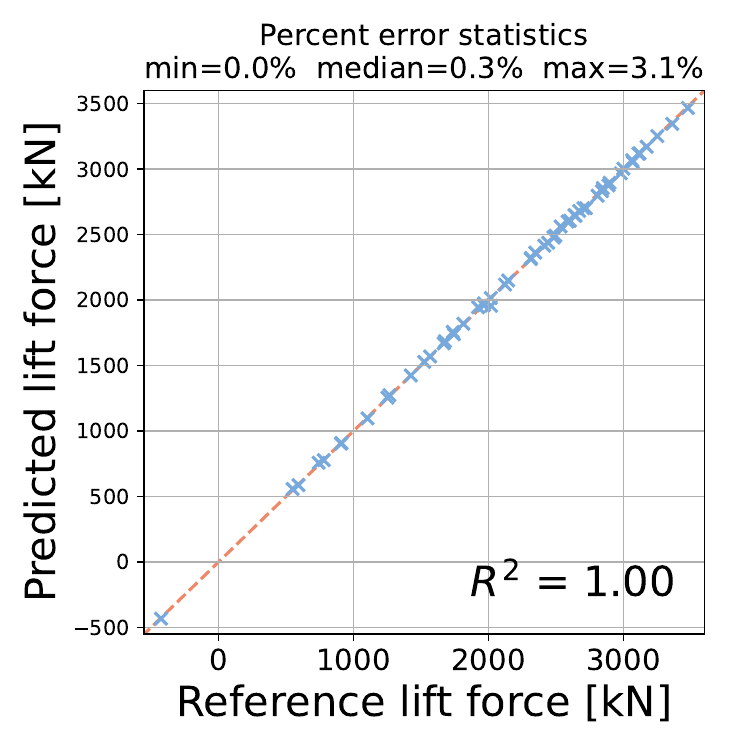}
\caption{224 samples}
\end{subfigure}%
\hfill
\begin{subfigure}{0.19\textwidth}
\centering
\includegraphics[width=\linewidth]{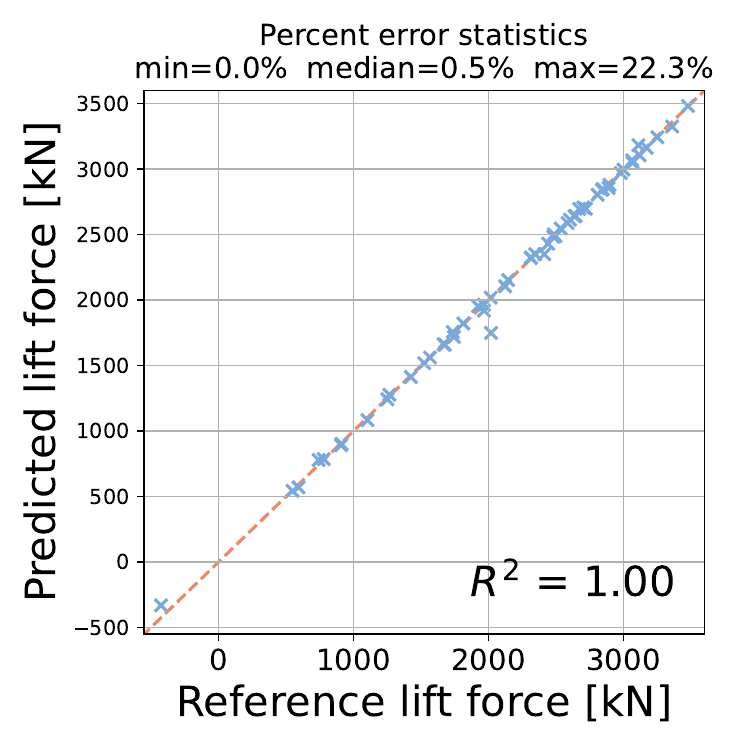}
\caption{112 samples}
\end{subfigure}%
\hfill
\begin{subfigure}{0.19\textwidth}
\centering
\includegraphics[width=\linewidth]{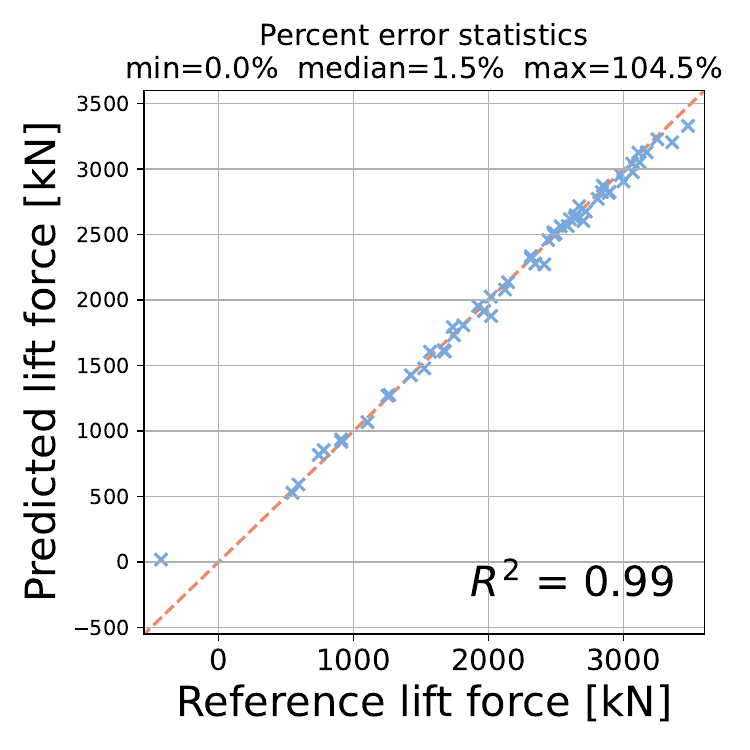}
\caption{56 samples}
\end{subfigure}%
\hfill
\begin{subfigure}{0.19\textwidth}
\centering
\includegraphics[width=\linewidth]{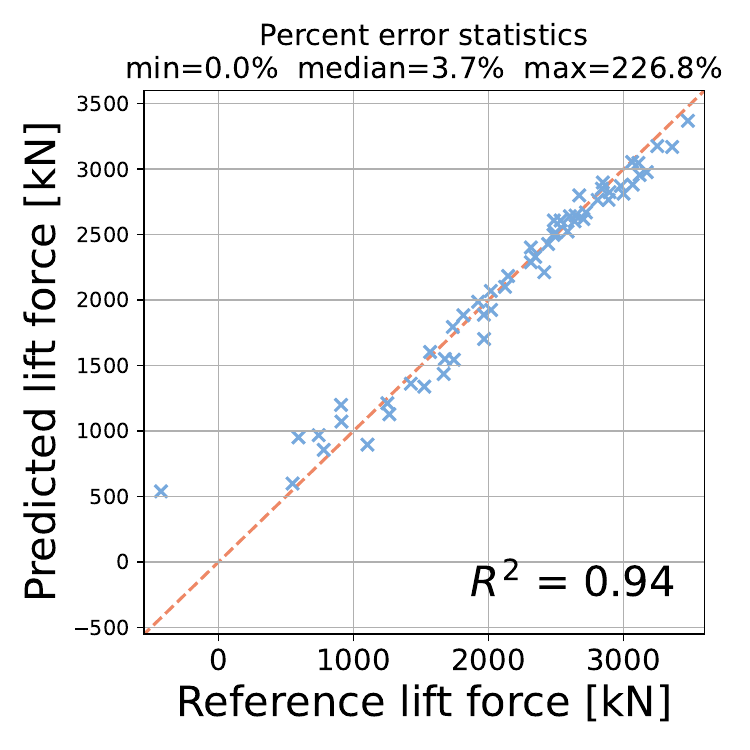}
\caption{28 samples}
\end{subfigure}%
\hfill
\begin{subfigure}{0.19\textwidth}
\centering
\includegraphics[width=\linewidth]{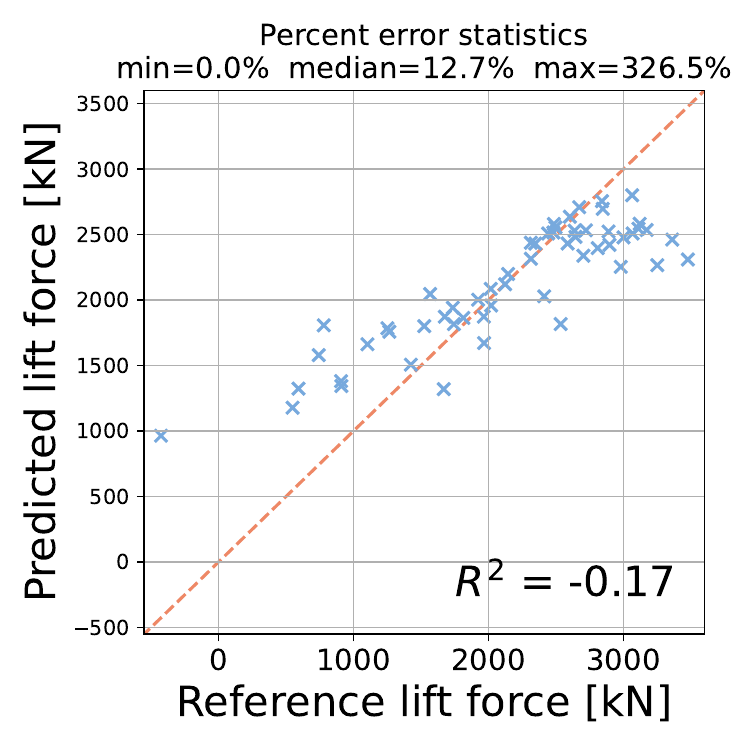}
\caption{14 samples}
\end{subfigure}
\caption{ Aerodynamic lift forces of Mach 0.85 test simulations when training on subsets. The full dataset (Figure~\ref{fig:mach085_lift_force}) contains 448 training samples. While median and maximum errors (as shown in the title of each plot) steadily increase, high accuracy is obtained with as little as 56 samples.}
\label{fig:wing_coefficients_diagonal_subsets}
\end{figure}

As we observe perfect $R^2$ correlation, we investigate training with fewer training simulations to see how many numerical simulations would be necessary to obtain a good surrogate model. In Figure~\ref{fig:wing_coefficients_diagonal_subsets}, we show aerodynamic lift force predictions for models trained on less data of the Mach 0.85 subset, showing strong performances even on little data.

\subsubsection{Wing pressure profiles}

\begin{figure}[h!]
\centering
\begin{subfigure}{0.3\textwidth}
\centering
\includegraphics[width=\linewidth]{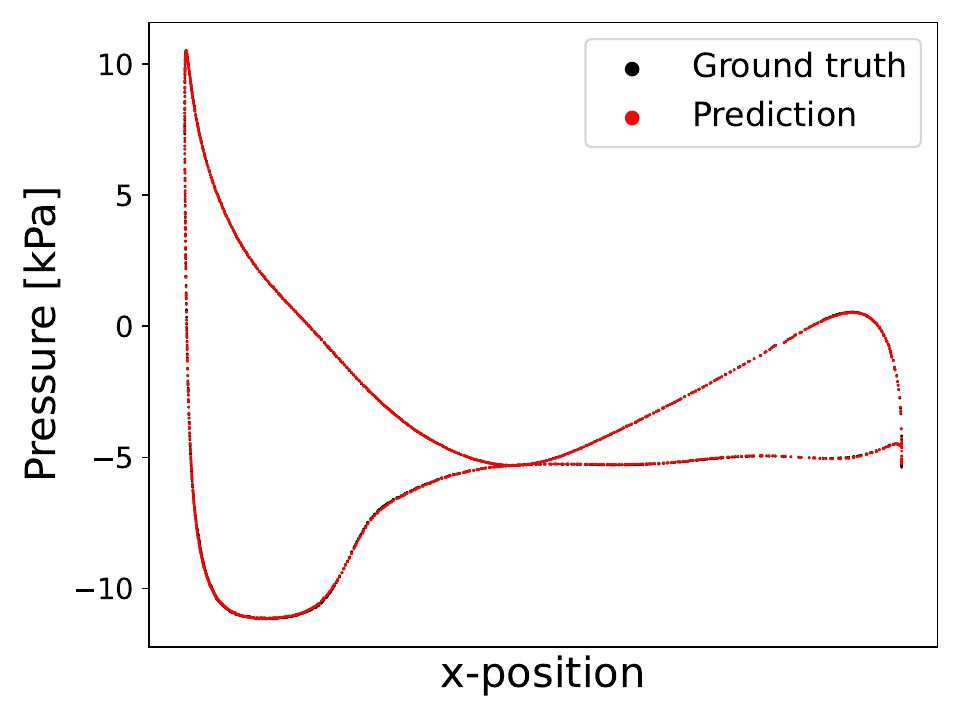}
\end{subfigure}%
\hfill
\begin{subfigure}{0.3\textwidth}
\centering
\includegraphics[width=\linewidth]{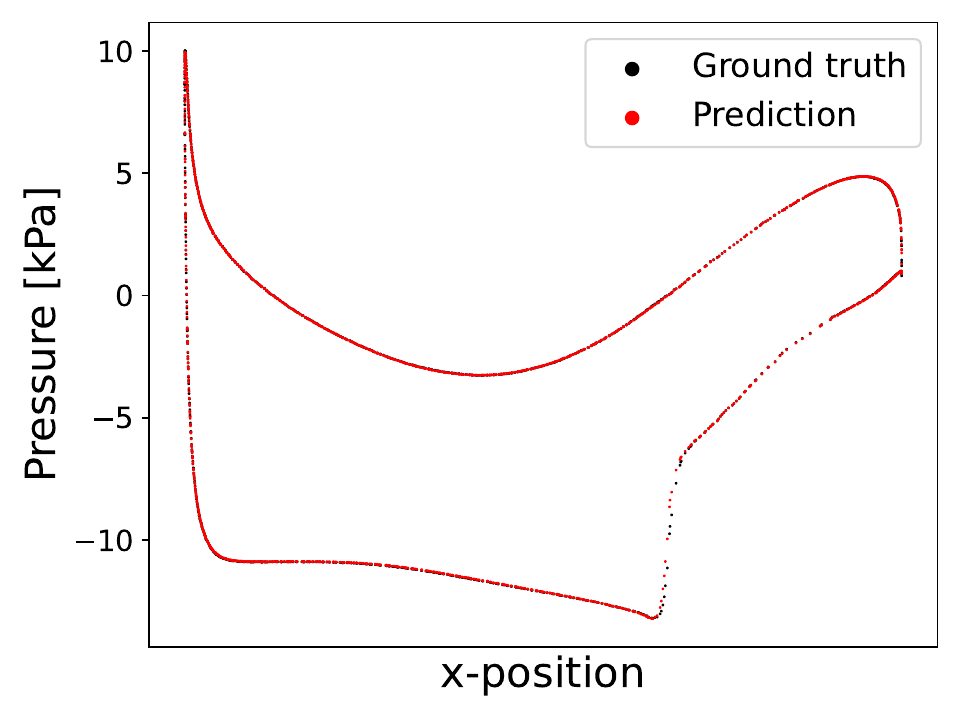}
\end{subfigure}%
\hfill
\begin{subfigure}{0.3\textwidth}
\centering
\includegraphics[width=\linewidth]{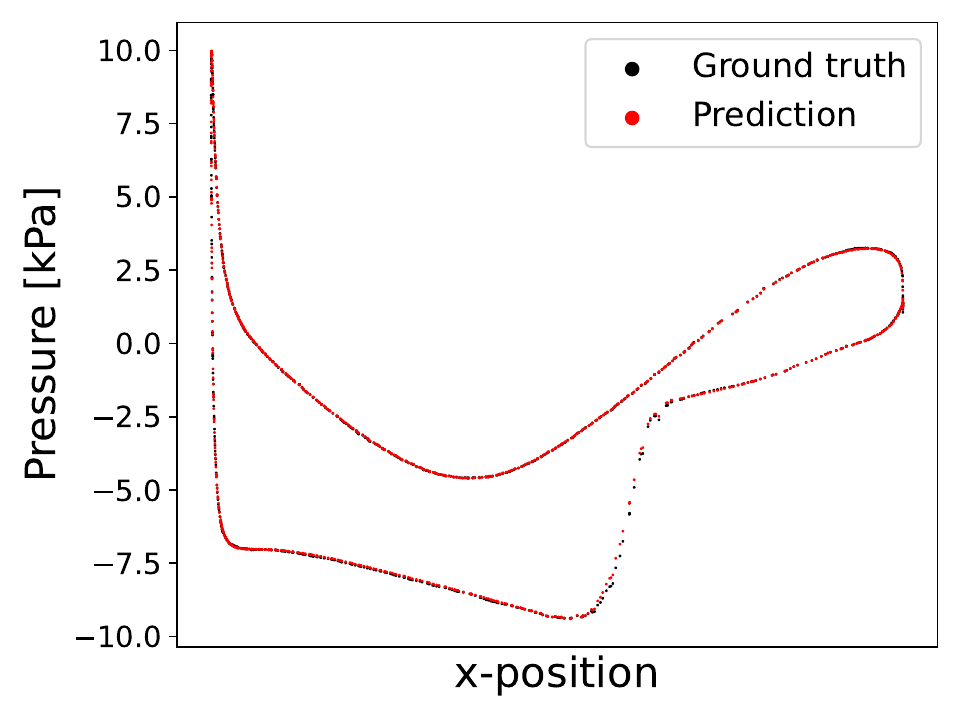}
\end{subfigure}
\begin{subfigure}{0.3\textwidth}
\centering
\includegraphics[width=\linewidth]{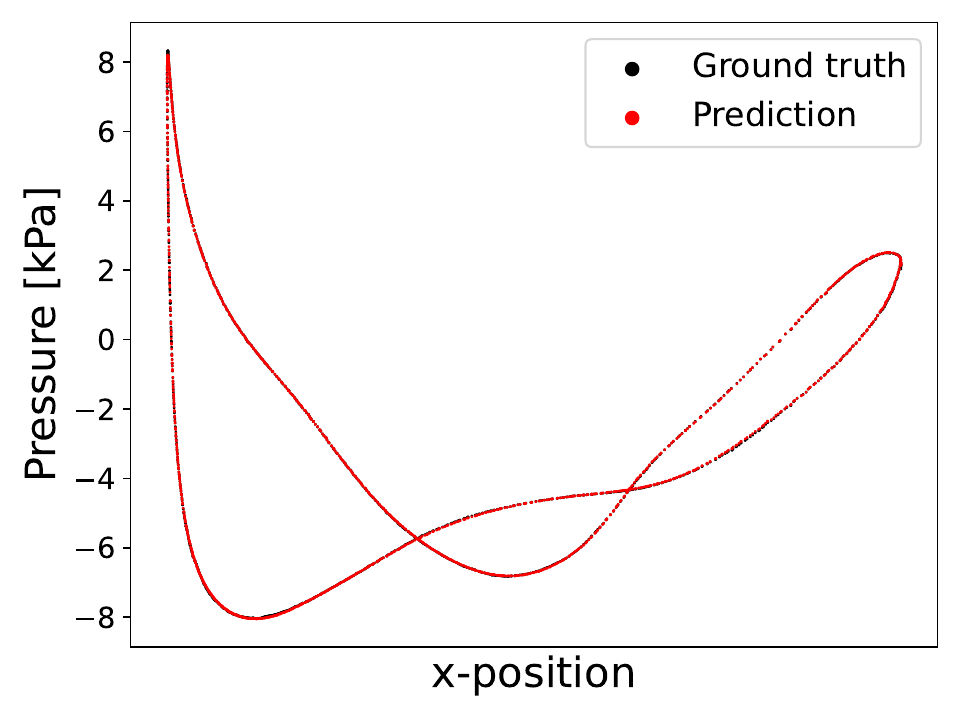}
\end{subfigure}%
\hfill
\begin{subfigure}{0.3\textwidth}
\centering
\includegraphics[width=\linewidth]{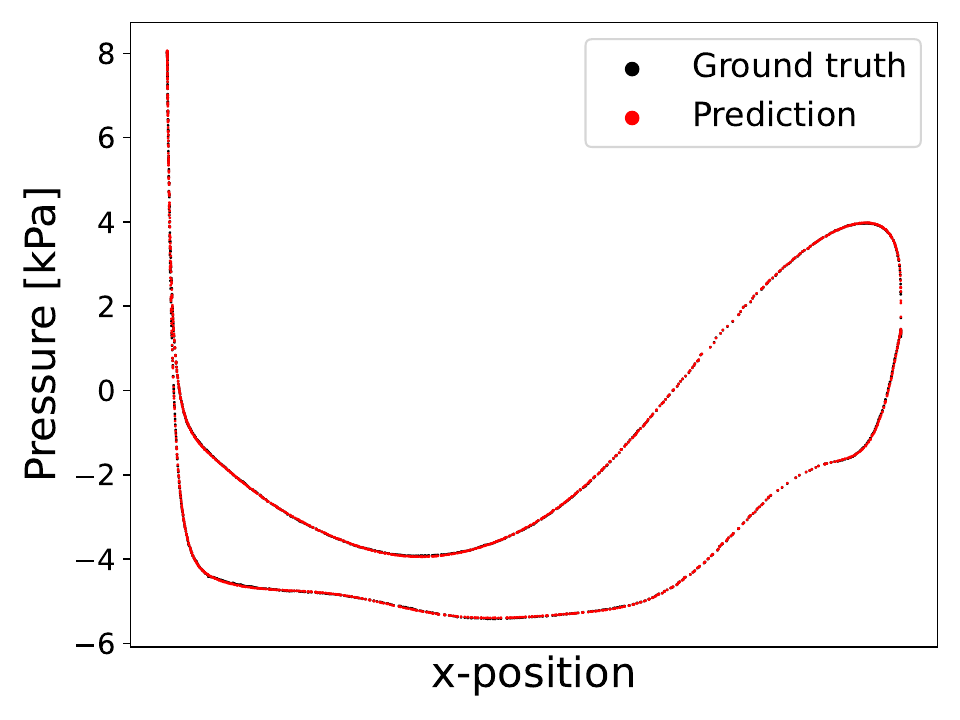}
\end{subfigure}%
\hfill
\begin{subfigure}{0.3\textwidth}
\centering
\includegraphics[width=\linewidth]{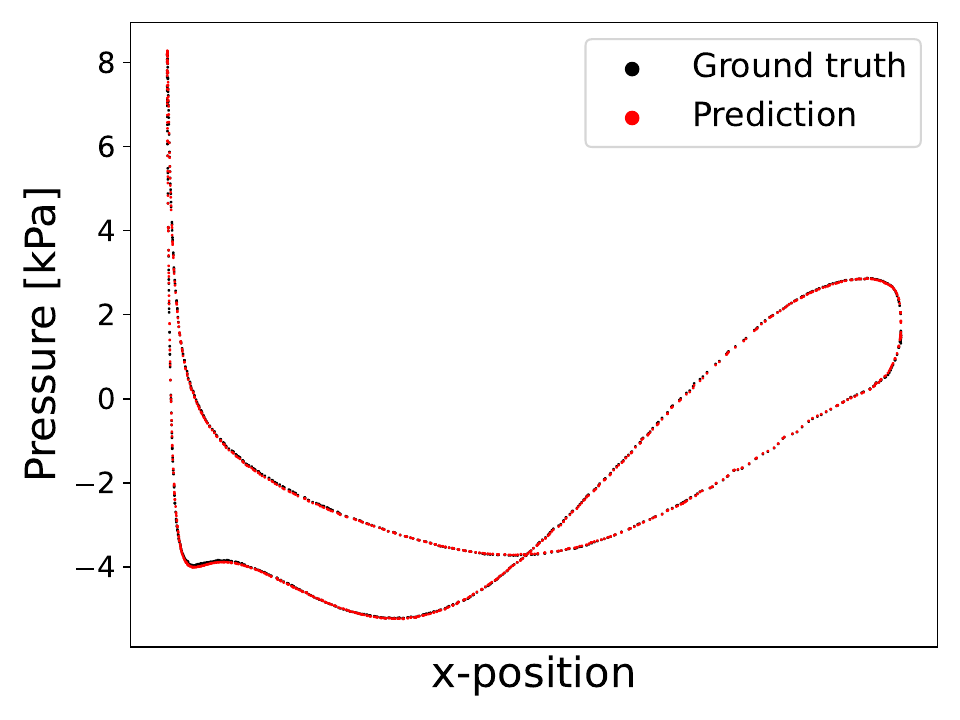}
\end{subfigure}
\begin{subfigure}{0.3\textwidth}
\centering
\includegraphics[width=\linewidth]{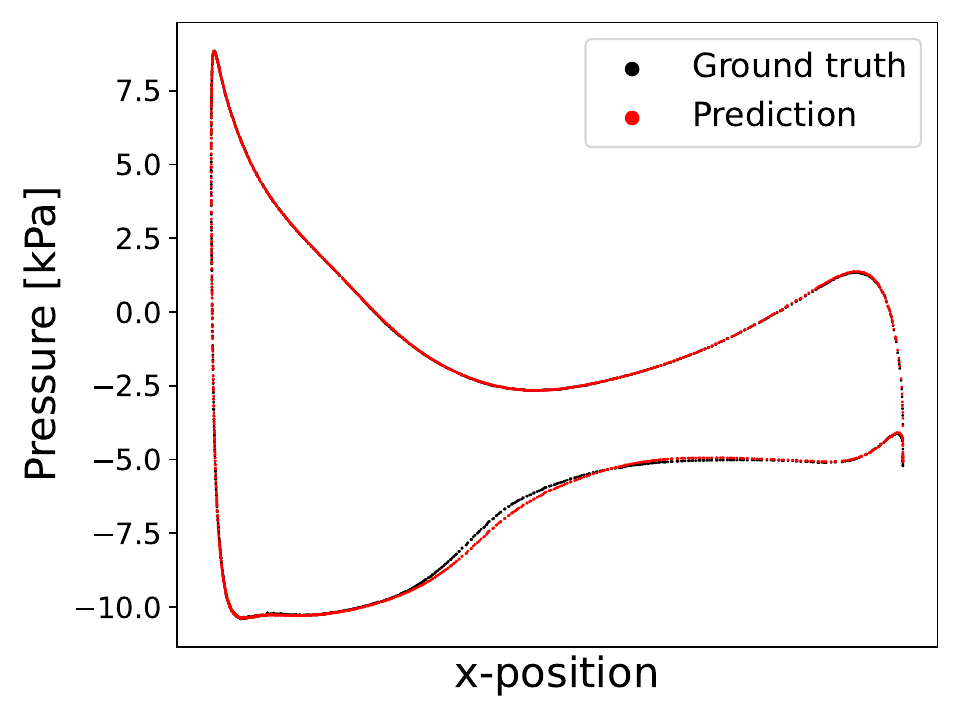}
\caption{15\% half-span}
\end{subfigure}%
\hfill
\begin{subfigure}{0.3\textwidth}
\centering
\includegraphics[width=\linewidth]{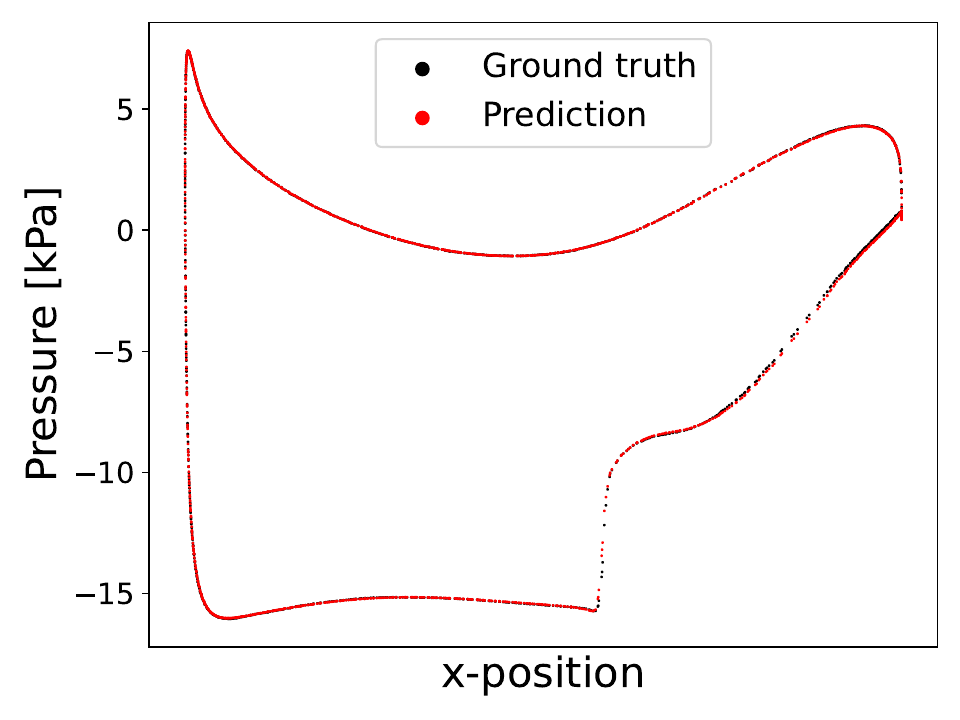}
\caption{50\% half-span}
\end{subfigure}%
\hfill
\begin{subfigure}{0.3\textwidth}
\centering
\includegraphics[width=\linewidth]{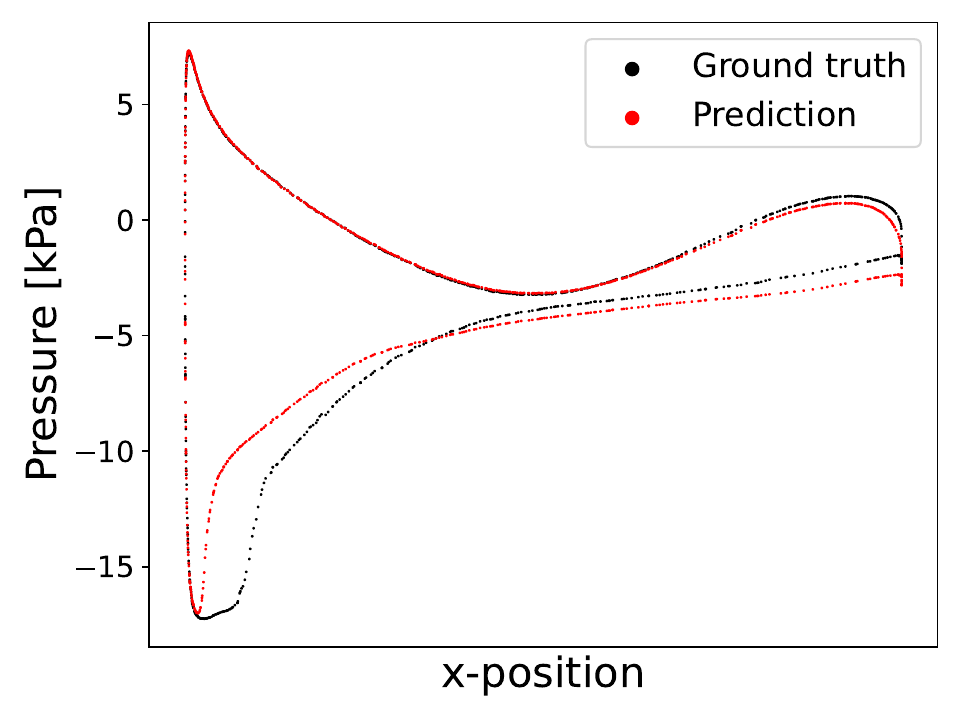}
\caption{95\% half-span}
\end{subfigure}
\caption{ Wing pressure profile visualization from the best (top), median (middle) and worst (bottom) prediction over all Mach 0.85 simulations from the testset. Slices are taken along the y-axis at 15\%, 50\% and 95\% of one side of the wing. AB-UPT obtains near perfect accuracy where even the worst-case obtains good pressure profiles in two out of three slices. }
\label{fig:wing_profiles}
\end{figure}

A common analysis in aerodynamic modeling is to visualize the pressure values along a slice in the wing. In Figure~\ref{fig:wing_profiles}, we show such wing pressure profiles at various slice locations in the wing, highlighting that AB-UPT can accurately reproduce such analysis results.

\subsubsection{Qualitative visualization}

\begin{figure}[h!]
\centering
\begin{subfigure}{0.3\textwidth}
    \centering
    \includegraphics[width=0.8\linewidth]{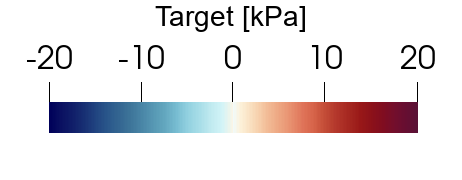}
\end{subfigure}%
\hfill
\begin{subfigure}{0.3\textwidth}
    \centering
    \includegraphics[width=0.8\linewidth]{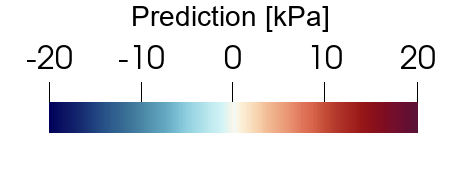}
\end{subfigure}%
\hfill
\begin{subfigure}{0.3\textwidth}
    \centering
    \includegraphics[width=0.8\linewidth]{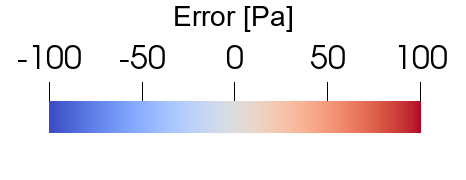}
\end{subfigure}
\begin{subfigure}{0.3\textwidth}
    \centering
    \includegraphics[width=\linewidth]{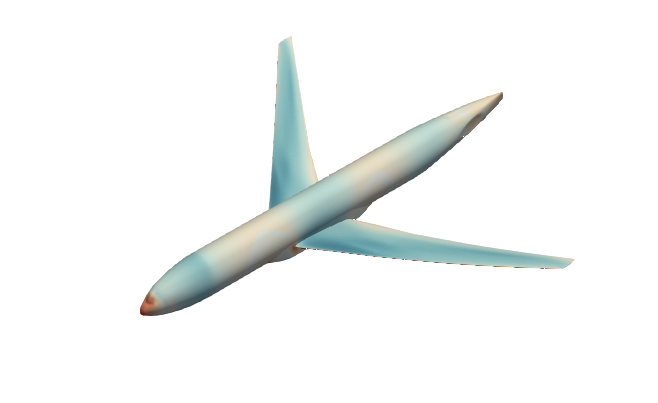}
\end{subfigure}%
\hfill
\begin{subfigure}{0.3\textwidth}
    \centering
    \includegraphics[width=\linewidth]{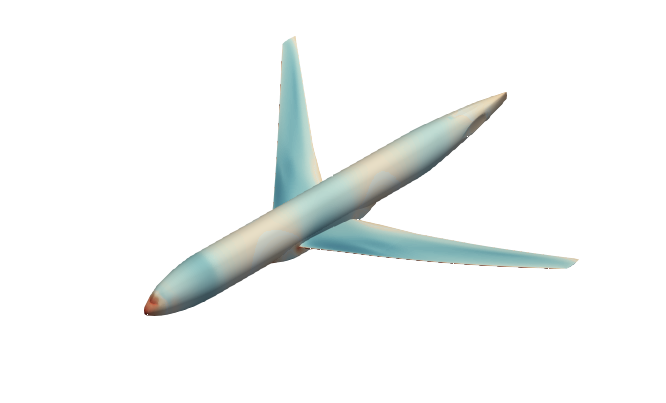}
\end{subfigure}%
\hfill
\begin{subfigure}{0.3\textwidth}
    \centering
    \includegraphics[width=\linewidth]{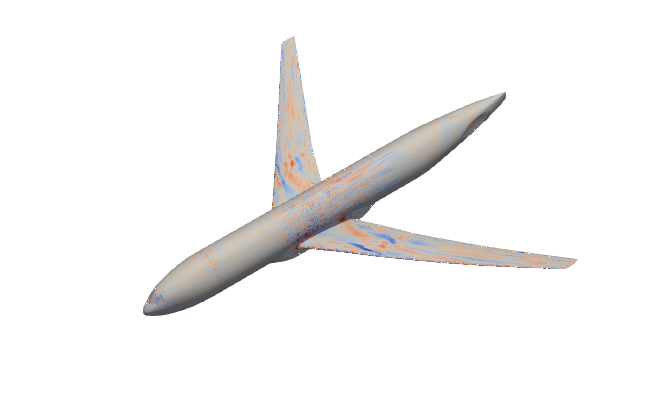}
\end{subfigure}
\begin{subfigure}{0.3\textwidth}
    \centering
    \includegraphics[width=\linewidth]{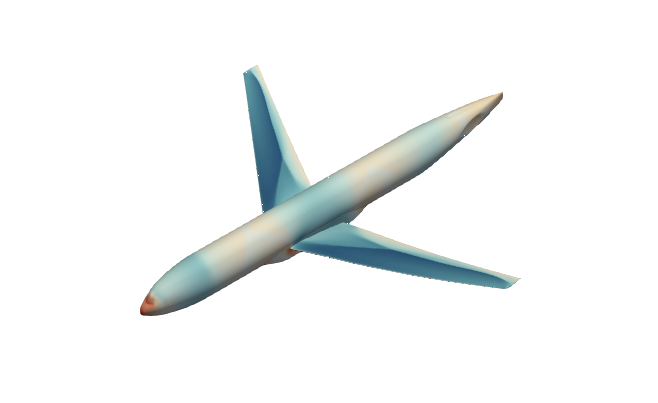}
\end{subfigure}%
\hfill
\begin{subfigure}{0.3\textwidth}
    \centering
    \includegraphics[width=\linewidth]{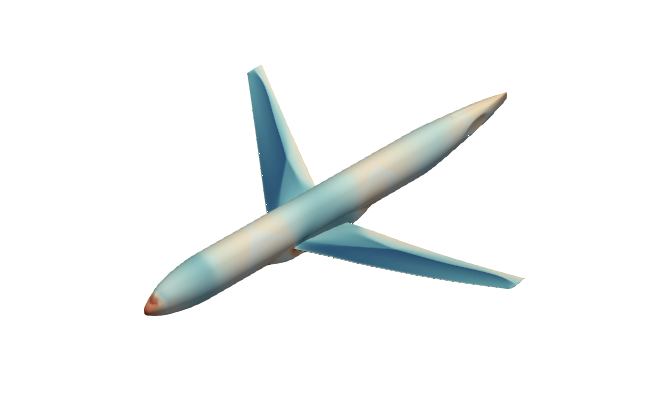}
\end{subfigure}%
\hfill
\begin{subfigure}{0.3\textwidth}
    \centering
    \includegraphics[width=\linewidth]{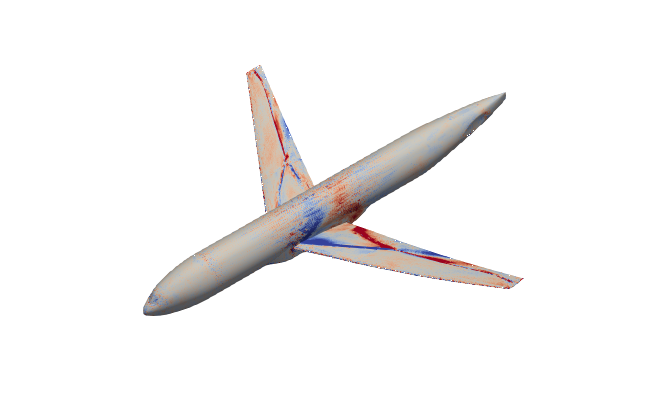}
\end{subfigure}
\begin{subfigure}{0.3\textwidth}
    \centering
    \includegraphics[width=\linewidth]{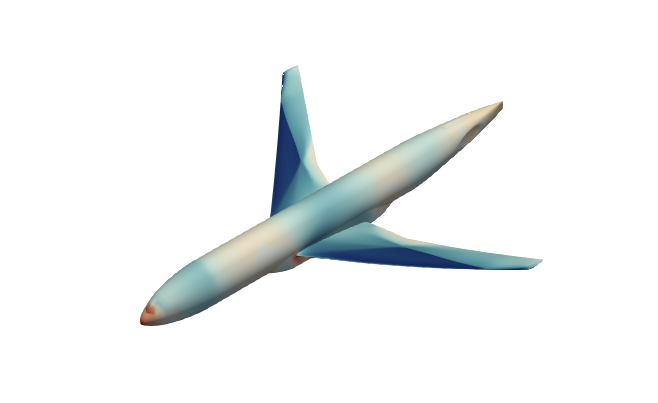}
\end{subfigure}%
\hfill
\begin{subfigure}{0.3\textwidth}
    \centering
    \includegraphics[width=\linewidth]{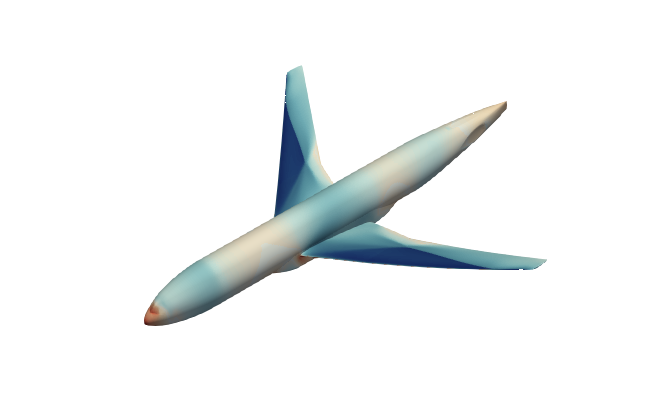}
\end{subfigure}%
\hfill
\begin{subfigure}{0.3\textwidth}
    \centering
    \includegraphics[width=\linewidth]{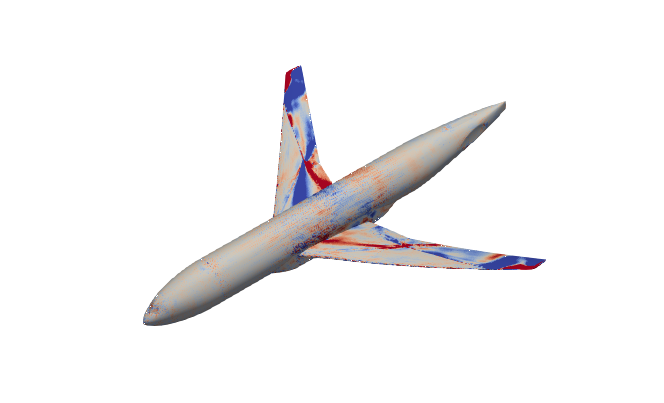}
\end{subfigure}
\caption{ Qualitative visualization of the surface pressure from the best (top), median (middle) and worst (bottom) prediction over all Mach 0.85 simulations from the testset. Note that the colorbar of the error was chosen relatively strict to highlight the errors where Pa is used as unit instead of kPa.  Pressure values are relative to the atmospheric pressure.
}
\label{fig:wing_visualization}
\end{figure}

\begin{figure}[h!]
\centering
\begin{subfigure}{0.3\textwidth}
    \centering
    \includegraphics[width=0.8\linewidth]{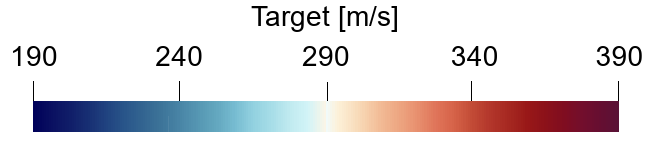}
\end{subfigure}%
\hfill
\begin{subfigure}{0.3\textwidth}
    \centering
    \includegraphics[width=0.8\linewidth]{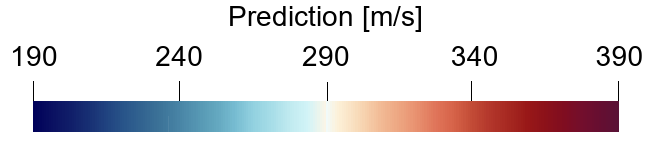}
\end{subfigure}%
\hfill
\begin{subfigure}{0.3\textwidth}
    \centering
    \includegraphics[width=0.8\linewidth]{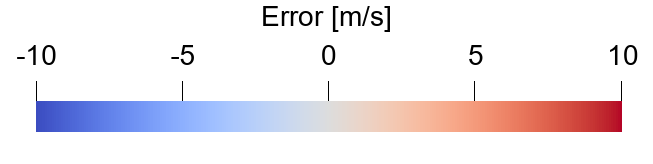}
\end{subfigure}
\begin{subfigure}{0.3\textwidth}
    \centering
    \includegraphics[width=\linewidth]{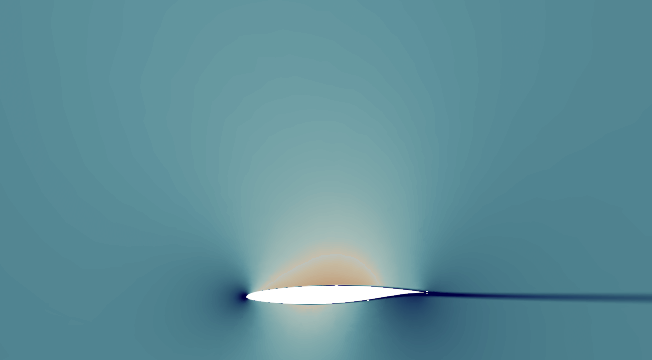}
\end{subfigure}%
\hfill
\begin{subfigure}{0.3\textwidth}
    \centering
    \includegraphics[width=\linewidth]{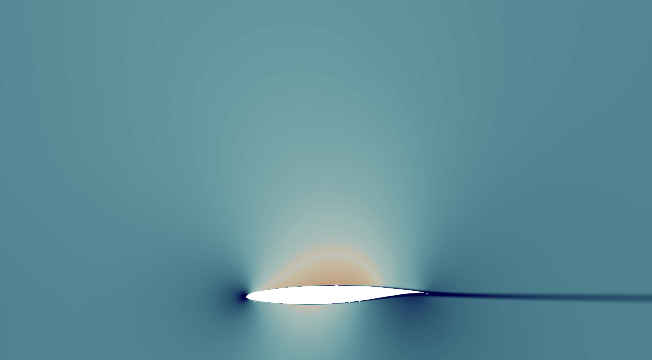}
\end{subfigure}%
\hfill
\begin{subfigure}{0.3\textwidth}
    \centering
    \includegraphics[width=\linewidth]{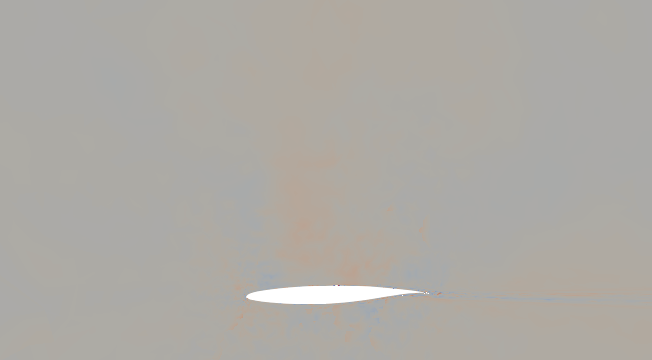}
\end{subfigure}
\begin{subfigure}{0.3\textwidth}
    \centering
    \includegraphics[width=\linewidth]{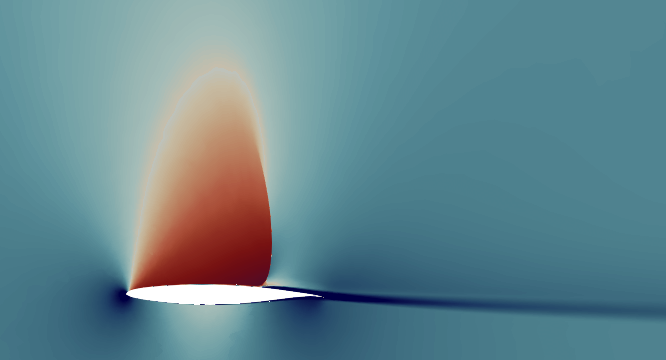}
\end{subfigure}%
\hfill
\begin{subfigure}{0.3\textwidth}
    \centering
    \includegraphics[width=\linewidth]{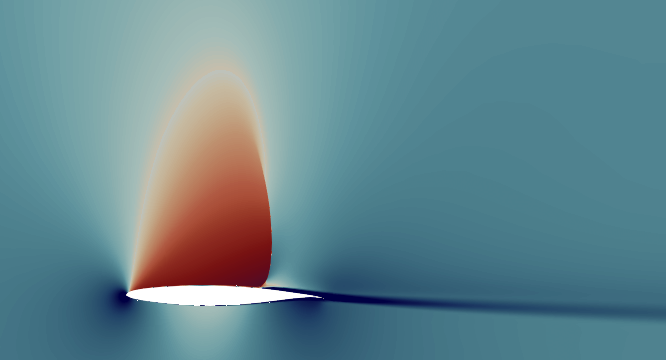}
\end{subfigure}%
\hfill
\begin{subfigure}{0.3\textwidth}
    \centering
    \includegraphics[width=\linewidth]{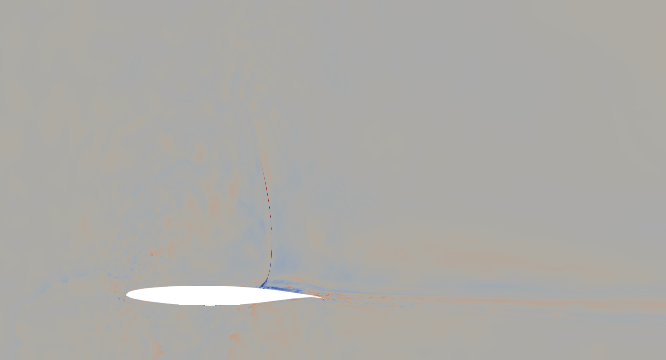}
\end{subfigure}
\begin{subfigure}{0.3\textwidth}
    \centering
    \includegraphics[width=\linewidth]{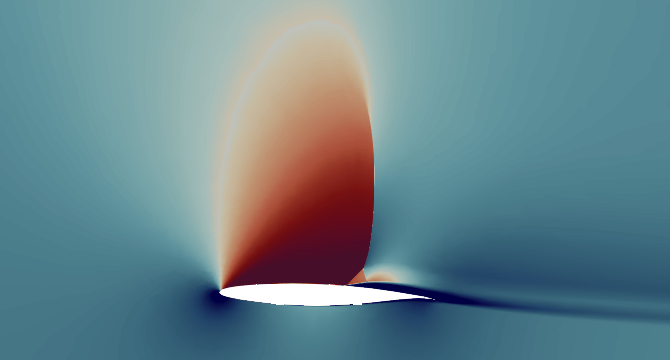}
\end{subfigure}%
\hfill
\begin{subfigure}{0.3\textwidth}
    \centering
    \includegraphics[width=\linewidth]{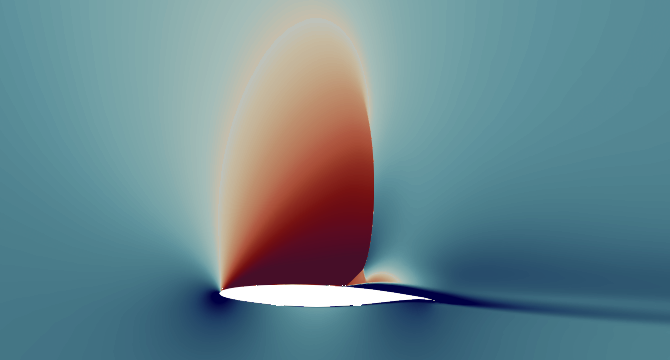}
\end{subfigure}%
\hfill
\begin{subfigure}{0.3\textwidth}
    \centering
    \includegraphics[width=\linewidth]{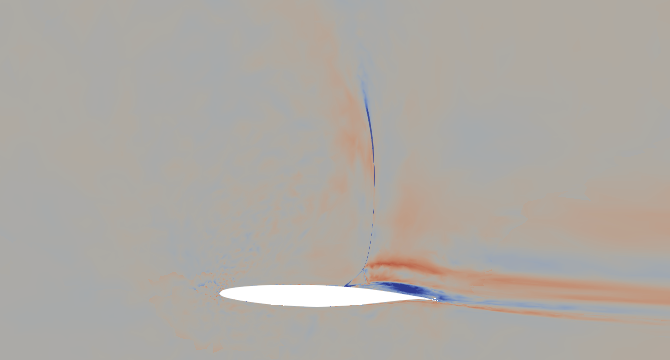}
\end{subfigure}
\caption{ Qualitative visualization of the x component of the velocity slice at 50\% half-span from the best (top), median (middle) and worst (bottom) prediction over all Mach 0.85 simulations from the testset. Due to geometric variations, the wing position varies in this visualization.
}
\label{fig:wing_volume_visualization}
\end{figure}

AB-UPT predictions are qualitatively visualized in Figure~\ref{fig:wing_visualization} and Figure~\ref{fig:wing_volume_visualization} showing high agreement with the numerical simulation result.




\subsubsection{Inference from isotropic geometry mesh}

\begin{figure}[h!]
\centering
\begin{subfigure}{0.5\textwidth}
\centering
\includegraphics[width=0.99\linewidth]{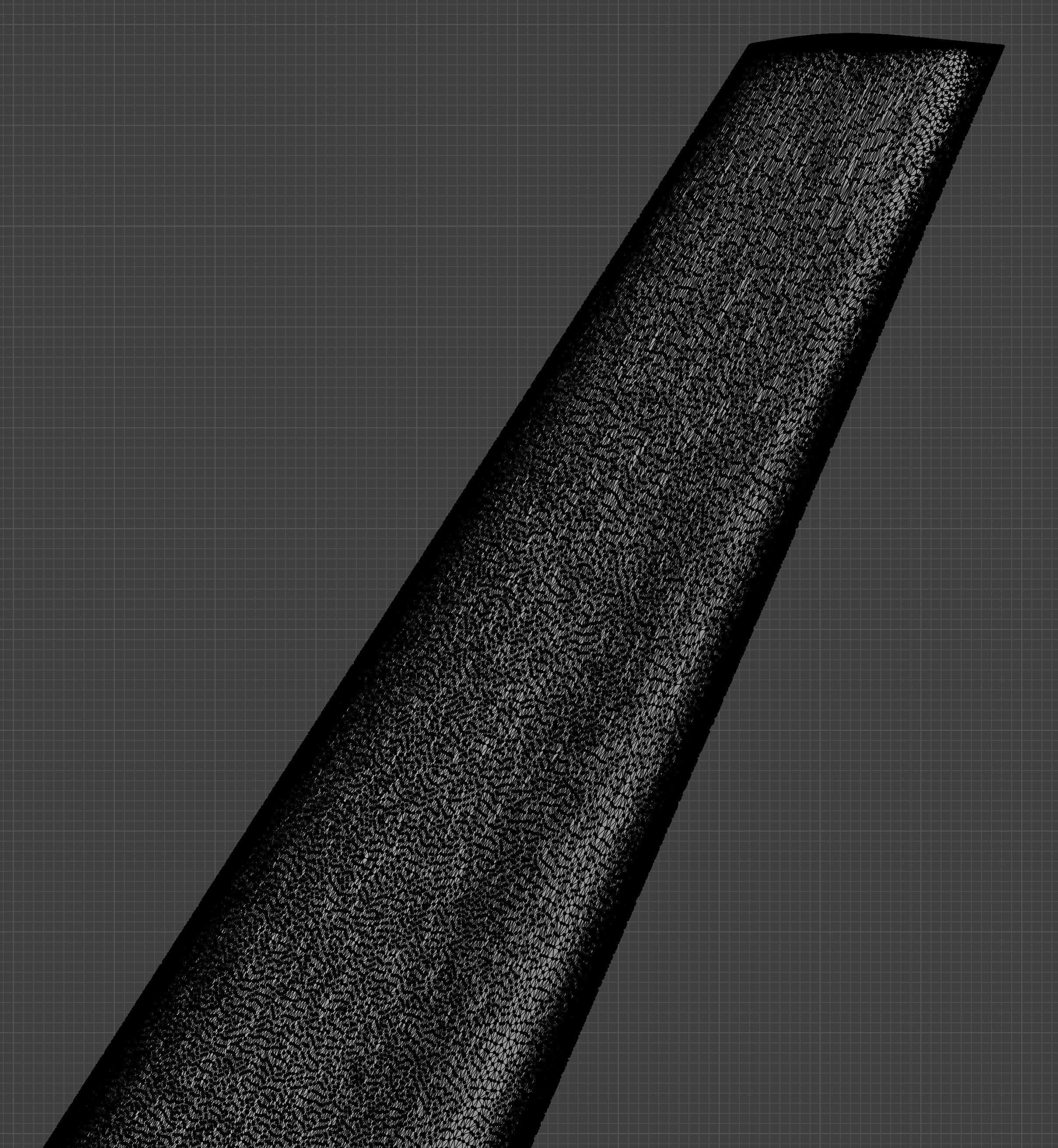}
\caption{Solution-adapted: 3M nodes}
\end{subfigure}%
\hfill
\begin{subfigure}{0.5\textwidth}
\centering
\includegraphics[width=0.99\linewidth]{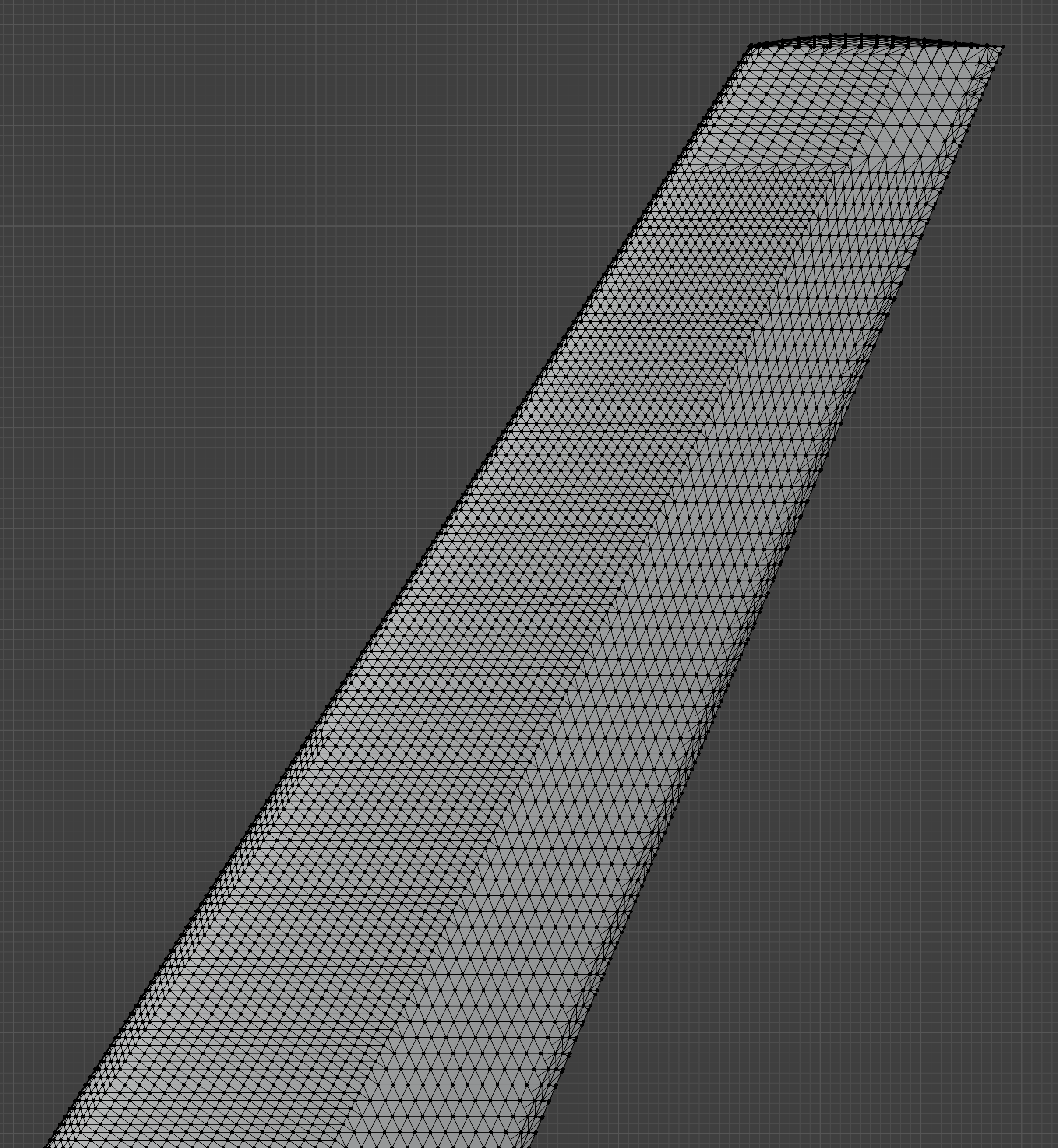}
\caption{Uniform: 200k nodes}
\end{subfigure}%
\caption{ Surface discretizations node distribution for a Mach 0.85 sample of SHIFT-Wing. Note the geometries of the wing are identical, simply the node distributions (their local densities) are different.}
\label{fig:wing_surface_densities}
\end{figure}

\begin{figure}[h!]
\centering
\begin{subfigure}{0.24\textwidth}
\centering
\includegraphics[width=\linewidth]{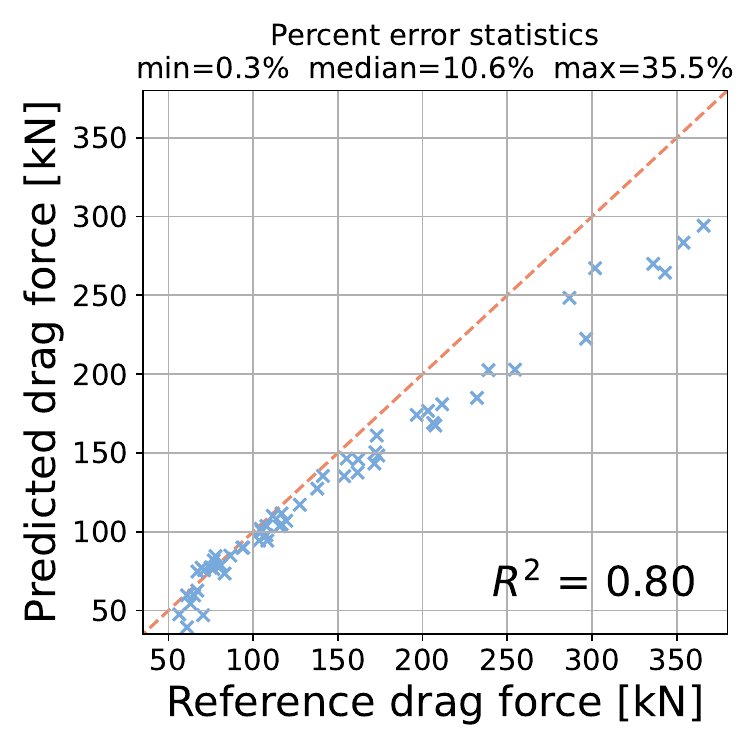}
\caption{``zero-shot'' geometry transfer (drag force)}
\end{subfigure}%
\hfill
\begin{subfigure}{0.24\textwidth}
\centering
\includegraphics[width=\linewidth]{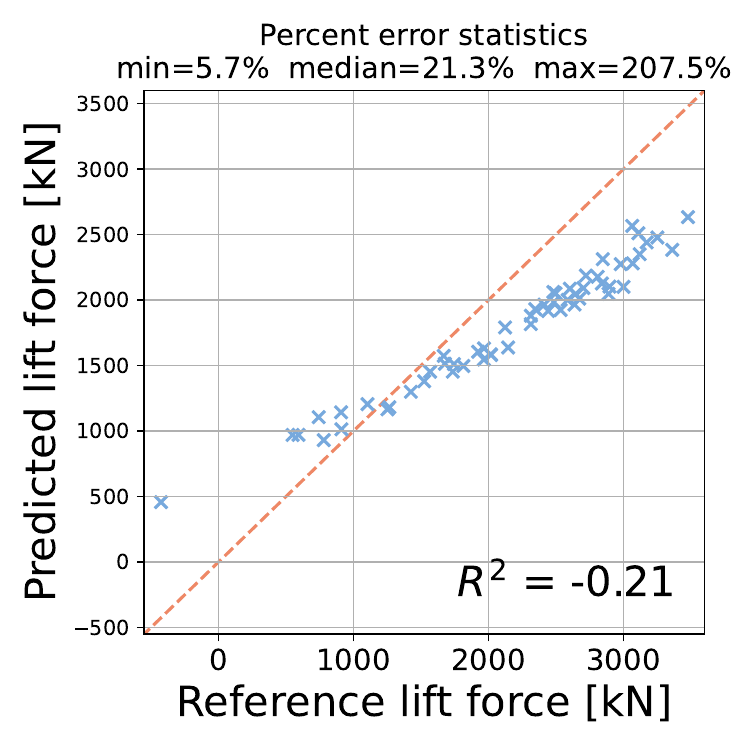}
\caption{``zero-shot'' geometry transfer 
(lift force)}
\end{subfigure}%
\hfill
\begin{subfigure}{0.24\textwidth}
\centering
\includegraphics[width=\linewidth]{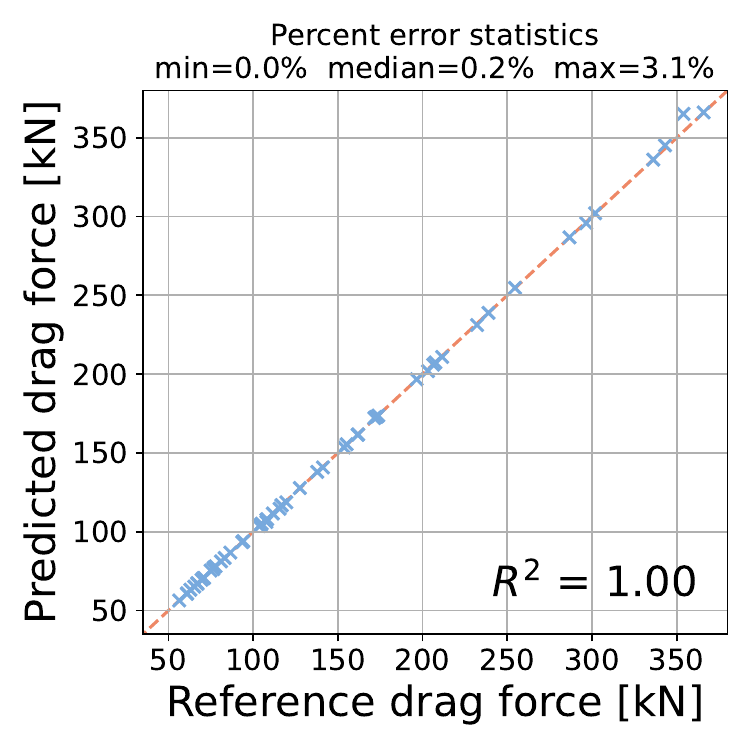}
\caption{Training with CAD inputs (drag force)}
\end{subfigure}%
\hfill
\begin{subfigure}{0.24\textwidth}
\centering
\includegraphics[width=\linewidth]{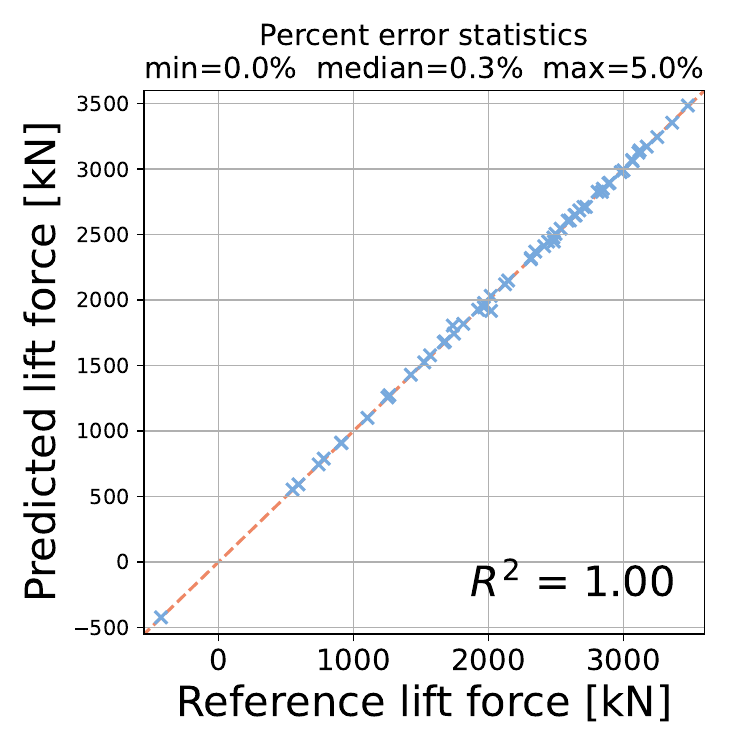}
\caption{Training with CAD inputs (lift force)}
\end{subfigure}
\caption{ Aerodynamic drag and lift forces from isotropic geometry for Mach 0.85 of the SHIFT-Wing dataset. (a-b) Transferring models in ``zero-shot'' fashion to the new geometry representation heavily degrades performance. (c-d) AB-UPT offers the flexibility to train models that can readily use a isotropic geometry by using the solution mesh only for query tokens. With this training procedure, AB-UPT obtains strong performances while requiring only a simple isotopically tessellated CAD geometry representation.}
\label{fig:wing_coefficients_diagonal_stl}
\end{figure}

The numerical solving process used to obtain the numerical solution for the airplane simulations employs Luminary Mesh Adaption, which ensures thin and sharp features, such as transonic shocks, are accurately captured by the solver with no user input and captures the wide range of flow conditions and geometries. As transformer-based surrogate models are commonly trained to map input points to target values, they use the solution mesh which underwent potentially multiple adaptation steps resulting in a highly anisotropic mesh. In contrast, in practice the solution is not known in advance and a user will typically provide a non-adapted geometry tessellation which is approximately isotropic. For neural surrogate models, this poses a shift in the input distribution as illustrated in Figure~\ref{fig:wing_surface_densities}. To investigate the impact thereof, we first evaluate performance of a model trained on the solution mesh and provide only the isotropic geometry mesh during inference. Figure~\ref{fig:wing_coefficients_diagonal_stl} (a-b) shows that simply plugging in the isotropic mesh into a model trained on the solution mesh (``zero-shot'' geometry transfer) greatly degrades performance.

To solve this issue, we exploit the flexibility of AB-UPT to decouple the point distribution of anchors and queries to allow the model to make predictions from an isotopically tessellated geometry. For this experiment, we sample (i) surface anchors and geometry branch inputs from the isotropic geometry mesh (ii) surface queries from the CFD solution mesh (iii) volume anchors from a regular grid in the domain (iv) volume queries from the CFD solution mesh. As target values are only available for the CFD solution mesh, we do not employ a loss onto the anchor tokens, but only on the query tokens. This setup teaches the model to only rely on isotropic geometry representation and regular grid volume representation as the positional information from the queries only influences the anchor representation via the backward pass but not via the forward pass. This setting is taken from the AB-UPT ``Training from CAD'' Section~\cite{alkin2025abupt}. Figure~\ref{fig:wing_coefficients_diagonal_stl} (c-d) shows that this setting is able to obtain strong performances without requiring the CFD solution mesh.

Note that in this setting, the reference aerodynamic forces are calculated from the solution mesh whereas the predicted forces are calculated from the isotropic CAD representation by calculating the corresponding surface normals and surface areas for Equation~\ref{eq:drag_and_lift_force}.


\subsubsection{Inference time analysis}

We present inference times of various settings in Table~\ref{table:wing_runtime}. For SHIFT-Wing, AB-UPT operates on 8K anchor tokens from which arbitrary many query tokens can be added. This provides fast inference in settings that do not require resolving the full CFD solution mesh. For example, inferring drag and lift forces from the CAD input with 200K points requires only 0.6 seconds as it is requires an order of magnitude less query tokens than decoding the full surface mesh.

\begin{table}[h!]
\centering
\begin{tabular}{lccc}
 & \#Surface points & \#Volume points & Runtime [s] \\
\midrule
Surface (isotropic CAD tesselation) & 200K & 8K & 0.6  \\
Surface (CFD solution mesh) & 3M & 8K & 8.4  \\
Full volume & 8K & 6M & 16.8 \\
Full volume + surface & 3M & 6M & 25.2 \\
\end{tabular}
\caption{ Approximate inference runtime to create predictions for various number of output points. The number of surface and volume anchors is fixed to 8K. Anchor points are always required as the model uses interleaved cross-attention between surface and volume anchors. }
\label{table:wing_runtime}
\end{table}

\section{Conclusion}

We provide an empirical evaluation of AB-UPT on two datasets generated with the Luminary SHIFT platform. SHIFT-SUV contains CFD simulations of different car shapes. Similarly, SHIFT-Wing contains airplane CFD simulations with varying geometry, angle of attack and Mach numbers. AB-UPT compares favorably against other neural surrogate models on these datasets, consolidating the notion of AB-UPT as the current state-of-the-art model for neural surrogate models in external CFD simulations. We show various qualitative and quantitative evaluations of our best model to show practical usability of the model. Notably, AB-UPT obtains perfect $R^2$ correlation of aerodynamic drag and lift forces, even when only provided with a isotopically tessellated geometry representation instead of an adaptively refined CFD mesh. Additionally AB-UPT is very fast to train, obtaining strong models in a day on a single GPU where aerodynamic forces of new designs can be calculated in as little as 0.6 seconds.

\section*{Acknowledgments}
We would like to thank Luminary Cloud for providing the high-quality SHIFT-SUV and SHIFT-Wing datasets. We extend our special gratitude to Michael Emory and Peter Lyu for their critical contributions to this work, which included detailed data descriptions, insightful visualizations, and fruitful discussions.

\newpage
\bibliographystyle{abbrvnat}
\bibliography{main}



\end{document}